\documentclass{article}

\usepackage{microtype}
\usepackage{graphicx}
\usepackage{subfigure}
\usepackage{booktabs} 

\usepackage{hyperref}
\usepackage{array} 
\usepackage{booktabs} 

\usepackage{wrapfig}

 \usepackage[accepted]{icml2024}

\usepackage{amsmath}
\usepackage{amssymb}
\usepackage{mathtools}
\usepackage{amsthm}

\usepackage[capitalize,noabbrev]{cleveref}

\theoremstyle{plain}

\theoremstyle{definition}

\theoremstyle{remark}

\usepackage[textsize=tiny]{todonotes}

\newcommand{\smallsec}[1]{\paragraph{#1.}}

\DeclareMathOperator*{\argmin}{arg\,min}

\icmltitlerunning{What is Dataset Distillation Learning?}

\begin{document}

\twocolumn[
\icmltitle{What is Dataset Distillation Learning?}



\icmlsetsymbol{equal}{*}

\begin{icmlauthorlist}
\icmlauthor{William Yang}{princeton}
\icmlauthor{Ye Zhu}{princeton}
\icmlauthor{Zhiwei Deng}{google}
\icmlauthor{Olga Russakovsky}{princeton}

\end{icmlauthorlist}

\icmlaffiliation{princeton}{Department of Computer Science, Princeton University, Princeton NJ, United States}
\icmlaffiliation{google}{Google Research, Mountain View CA, United States}

\icmlcorrespondingauthor{William Yang}{williamyang@cs.princeton.edu}

\icmlkeywords{Machine Learning, ICML}

\vskip 0.3in
]



\printAffiliationsAndNotice{}  

\begin{abstract}
Dataset distillation has emerged as a strategy to overcome the hurdles associated with large datasets by learning a compact set of synthetic data that retains essential information from the original dataset. While distilled data can be used to train high performing models, little is understood about how the information is stored. In this study, we posit and answer three questions about the behavior, representativeness, and point-wise information content of distilled data. We reveal distilled data cannot serve as a substitute for real data during training outside the standard evaluation setting for dataset distillation. Additionally, the distillation process retains high task performance by compressing information related to the early training dynamics of real models. Finally, we provide an framework for interpreting distilled data and reveal that individual distilled data points contain meaningful semantic information. This investigation sheds light on the intricate nature of distilled data, providing a better understanding on how they can be effectively utilized.
\end{abstract}

\section{Introduction}
The past decade in machine learning research has been marked with incredible breakthroughs in leveraging over-parameterized models. As a consequence, the landscape of machine learning research has been increasingly dominated by massive datasets. The growing size of training datasets presents a new challenge on the infrastructure required to store and train on such data. Large-scale data not only strains existing compute infrastructure with increased training times but also limits its accessibility to researchers with sufficient compute infrastructure. Therefore, there is a crucial need for scaling down large-scale datasets. 
\begin{figure}
    \centering
    \includegraphics[width=0.99\linewidth]{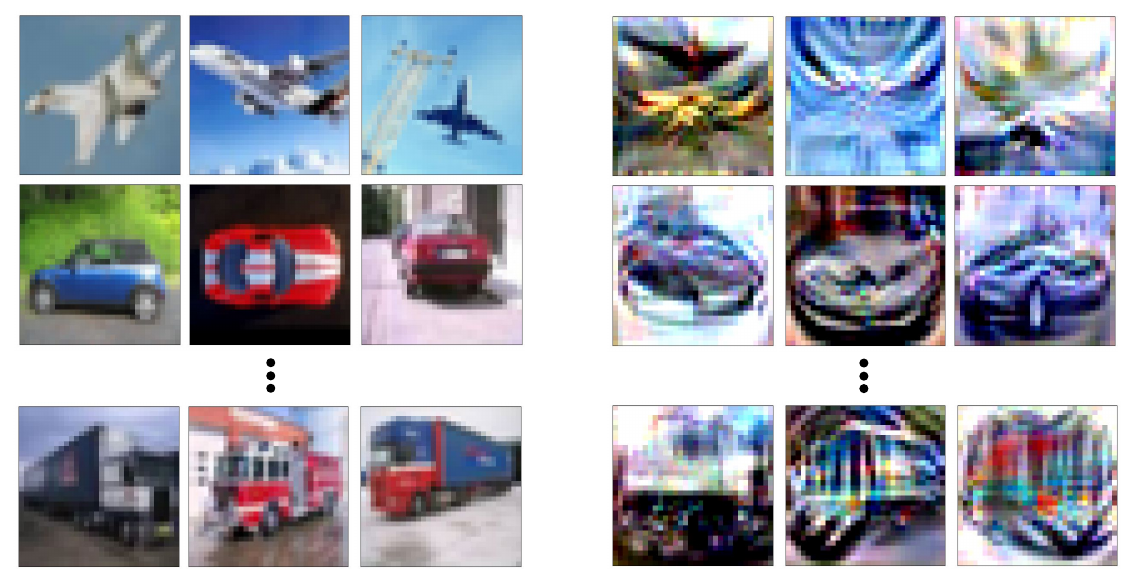}
    \caption{\textbf{Real vs. distilled data.} Real images of airplane, car, and truck from CIFAR-10 \cite{krizhevsky2009learning} are shown on left and highly salient distilled images of the same classes are shown on the right. While distilled images \emph{can} be used to train high-accuracy classifiers, \emph{why} this is possible and \emph{what} do they represent remains unclear.}
    \label{fig:hook}
\end{figure}

The crux of the problem lies in the ability to scale-down a large dataset without losing any essential information. Classical data compression algorithms attempt this by selecting the representative images in the training data \cite{guo2022deepcore}. However, such algorithms are limited by the finite number of data points that are present in the training dataset, which can be very restrictive in its representation. Dataset distillation overcomes this limitation by \emph{learning} a small, information-dense dataset that can serve as a substitute for the original dataset \cite{wang2018dataset, sachdeva2023data}. 

Learning a synthetic dataset can be a double-edged sword: on one hand, dataset distillation synthesizes a small, information dense dataset that outperforms classical data compression techniques \cite{sachdeva2023data}, but on the other hand, the distilled data does not look like real data (Figure \ref{fig:hook}) and can behave differently than real data \cite{zhong2023towards}. Therefore, it is very important to identify when and why distilled data fails to be an effective drop-in replacements for real data. 

Current literature on dataset distillation analysis is very sparse. \cite{vicol2022implicit} studies the implicit bias between warm start vs. cold start of the bilevel optimization under the meta-model matching dataset distillation approaches. \cite{schirrmeister2022less} shows that dataset distillation methods can be regularized towards a simpler dataset using a pre-trained generative model. \cite{maalouf2023size} provides theoretical support for the existence of small distilled dataset in the context of kernel ridge regression models. However, to the best of our knowledge, no work has been done to understand the information and dynamics of distilled data post-distillation process.

To gain a deeper understanding into dataset distillation, we posit and answer three important questions on the nature of distilled data.\footnote{\url{https://github.com/princetonvisualai/What-is-Dataset-Distillation-Learning}}

\textbf{To what extent can distilled data act as substitute for real data?} Dataset distillation is commonly motivated by synthesizing information rich data that can serve as effective drop-in replacements of the original dataset \cite{sachdeva2023data}. However, this comes with limitations and trade-offs. Prior works have already identified one such limitation: data distilled using one model architecture can \emph{not} be effectively used to train a different model architecture \cite{zhong2023towards}. We perform additional analyses to reveal the extent of distilled data's ability to substitute for real data. We demonstrate that models trained on real data are able to successfully recognize the classes in distilled data, demonstrating that distilled data does encode transferable semantics. However, at training time, simply mixing real data with distilled data results in \emph{decrease} in the performance of the final classifier. Therefore, distilled data should not be treated as real data during training, and we have to be careful in training with distilled data outside the typical evaluation setting for dataset distillation (training only on distilled data and on the same model architecture). 

\textbf{What kind of information is captured in distilled data?} While distilled data results in models that are able to classify real data, it is unclear what information is actually being stored. Our analyses suggest that distilled data captures the same information that would be learned from real data early in the training process. We demonstrate this from three perspectives. First, we reveal strong parallels in predictions between models trained on distilled data and models trained on real data with early stopping. Next, we uncover that a model trained on real data learns to recognize distilled data early in the training. Finally, we study the loss curvature of a model trained on real data with respect to distilled data, and show that the curvature induced by distilled data (for some distillation algorithms) decreases to low values quickly during training, indicating that distilled data captures little additional information beyond the early training.

\textbf{Do distilled data points individually carry meaningful information?} Given that the whole distilled dataset compresses the early training dynamics, are individual examples still meaningful? To answer this question, we introduce a new interpretable framework for distilled data by leveraging a popular interpretability method: influence functions \cite{koh2017understanding}. In contrast to their intended uses to better understand a model, we utilize influence functions to better understand the data. We empirically demonstrate the power and consistency of the framework and reveal that notable semantic information is stored in different distilled data points: for example, one distilled image is associated with classifying yellow cars whereas another with cars in parking lots. 

\section{Preliminary}
In this section, we set up the relevant background and training procedure for our analysis. Dataset distillation methods can be assorted into four categories \cite{sachdeva2023data}: meta-model matching \cite{nguyen2020dataset, nguyen2021dataset, deng2022remember, zhou2022dataset}, distribution matching \cite{ wang2022cafe, zhao2023dataset, zhao2023improved}, gradient matching \cite{zhao2021dataset, jiang2023delving}, and trajectory matching \cite{cazenavette2022dataset, cui2023scaling, wu2023multimodal}. We use a diverse set of methods by picking one common baseline for each category: (1) the meta-model learning matching algorithm Back-Propagation Through Time (BPTT) \cite{deng2022remember}, (2) distribution matching \cite{zhao2023dataset}, (3) gradient matching \cite{zhao2020dataset}, and (4) trajectory matching \cite{cazenavette2022dataset}.

Denoting $\mathcal{X}_s$ as the learnable synthetic data, $ \mathcal{Y}_s$ as fixed labels attributed to each distilled data point, $\mathcal{X}_r, \mathcal{Y}_r$ as the real dataset, and some model with parameters $\theta$ as $\mathcal{F}_\theta$, BPTT (1) performs the dataset distillation through bi-level optimization formulated as $\argmin_{\mathcal{X}_s} L(\mathcal{F}_{\theta_s}(\mathcal{X}_r), \mathcal{Y}_r)$ such that $\theta_s = \argmin_\theta L(\mathcal{F}_\theta(\mathcal{X}_s), \mathcal{Y}_s)$.

Distribution matching (2) also distills data at a class level and uses penultimate features from random convolutional neural networks as $\mathcal{F}^p$ and similarity function $D$ defined as the maximum mean discrepancy \cite{gretton2012kernel} to arrive at the objective $\argmin_{\mathcal{X}_s} D[\mathcal{F}^p(\mathcal{X}_s), \mathcal{F}^p(\mathcal{X}_r)]$.

Gradient matching (3) perform class-level optimization (only use synthetic and real images of matching classes) to match the gradients of model parameters $\theta$ given some similarity function $D$ as:  $\argmin_{\mathcal{X}_s}  D[\nabla L(\mathcal{F}_\theta(\mathcal{X}_s), \mathcal{Y}_s), \nabla L(\mathcal{F}_\theta(\mathcal{X}_r), \mathcal{Y}_r)].$

Finally, trajectory matching (4) directly matches a portion of the training trajectories by minimizing the objective $\frac{\|\hat{\theta}_{t+N} - \theta^*_{t+M}\|^2_2 }{\|\theta^*_t - \theta^*_{t+M}\|^2_2}$ where $\hat{\theta}_{t+N}$ is the model trained with distilled data for $N$ steps from $\theta^*_t$ and $\hat{\theta}_{t+M}$ is the model trained with real data for $M$ steps from $\theta^*_t$ ($N$, $M$, and $t$ are hyperparameters).

\begin{figure}
    \centering
    \includegraphics[width=0.99\linewidth]{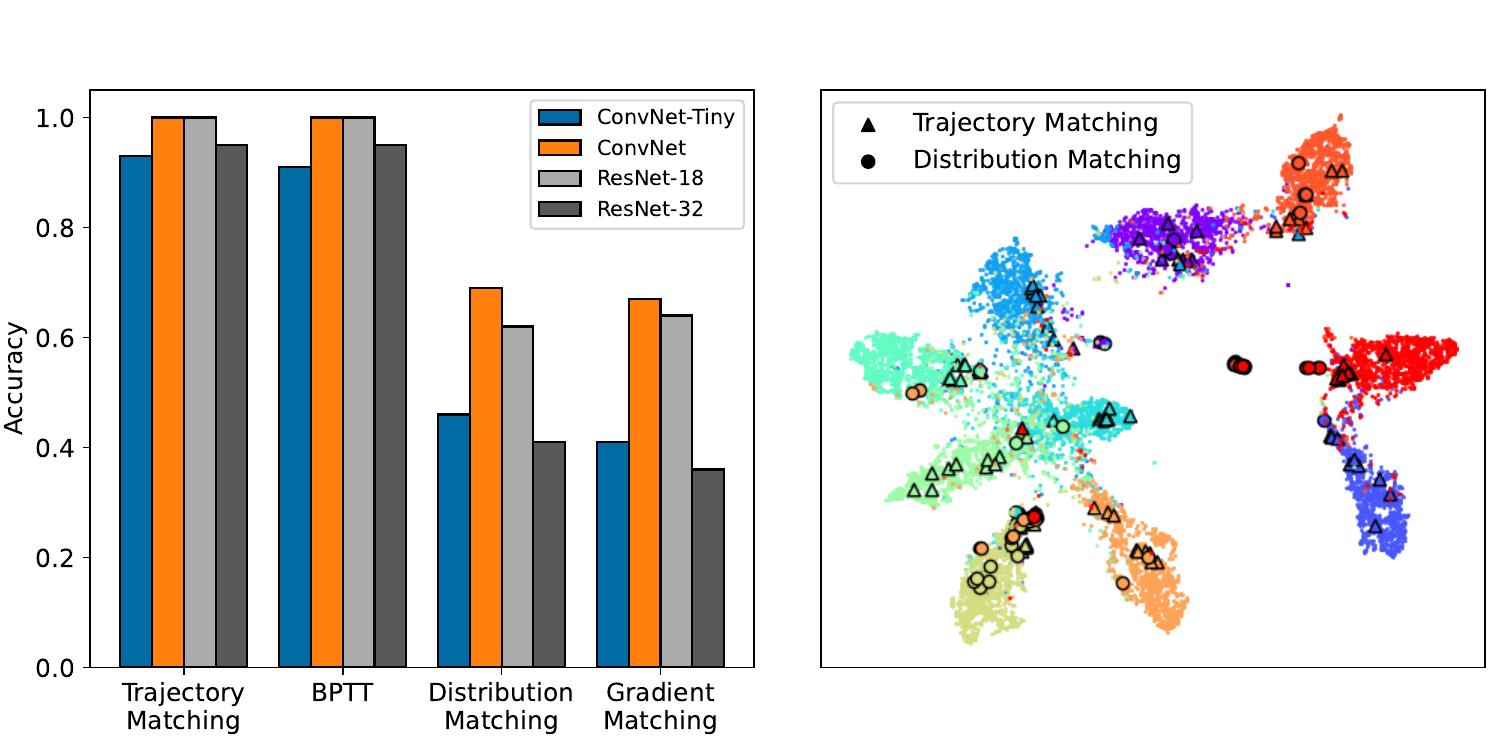}
    \caption{\textbf{Pre-trained models recognize distilled data.} \textit{left.} Classification accuracy of four different architectures (bar colors) trained on the real training dataset and evaluated on 100 images distilled using four different distillation algorithms (x-axis). These models successfully recognize distilled data (distribution matching and gradient matching do less well but they are known to distill less information than the other two). \textit{right.} UMAP \cite{mcinnes2018umap} visualization of real test images and distilled images using the penultimate features of a ResNet-18 \cite{he2016deep} model trained on real data. Most of the distilled images lie on the class clusters (indicated by the color), revealing that classification models do interpret distilled images similar to real images.}
    \label{fig:model}
\end{figure}

\smallsec{Experimental setup} We leverage the CIFAR-10 \cite{krizhevsky2009learning} dataset for our analysis with additional analyses on other datasets in Section \ref{sec:add_pred_analysis} of the appendix. We use the standard three layers deep, 128 filters wide convolutional neural networks to train on distilled data and real data with 0.01 learning rate and 0.9 momentum for 300 iterations using SGD optimizer. We leverage the distilled data provided by the authors for gradient matching (3) and trajectory matching (4) while we reproduce the distilled data for BPTT (1) and distribution matching (2). In contrast to the original BPTT paper, we distill images initialized from real images rather than uniform Xavier initialization \cite{glorot2010understanding} since it gives better behaved distilled data (details provided in section \ref{sec:bptt_init} of the appendix).  

\section{Distilled vs. Real Data}
\label{sec:distilledvsreal}
Having set up the preliminaries and experiments, we are ready to tackle the first question: \textit{To what extent does distilled data act like real data?} We show that distilled data is recognizable by models trained on real data, suggesting that  distilled data captures sufficient class semantics to be recognizable. However, distilled data does not appear to lie on the real data manifold and further, training on distilled data is very sensitive. Beyond the well-known result about the limited cross-architecture generalization of distilled data as opposed to real data~\cite{zhong2023towards}, we demonstrate that the distilled data and real data do not combine well during training -- in fact, adding real data samples to distilled data may \emph{decrease} the accuracy of the trained model. 

\smallsec{Distilled data (like real test data) is recognizable by models trained on real data} The standard dataset distillation pipeline involves training on distilled data and evaluating on real data. Not surprisingly, this performs well as this is explicitly optimized. However, is the distilled data capturing meaningful semantics of the real data? To check this, we setup the inverse pipeline: training on real images and evaluating on distilled images. The accuracy on the distilled data shown in Figure \ref{fig:model} \textit{left} reveals high classification accuracy on distilled data, which suggests that semantics in distilled data are transferable. Furthermore, to better understand how real-trained-models see distilled data, we interpret their inner workings by visualizing their penultimate features. Using UMAP \cite{mcinnes2018umap} on penultimate features of a ResNet-18 \cite{he2016deep} trained on real data of distilled data and real test data shown in Figure \ref{fig:model} \textit{right}, we observe that the distilled data appear to lie on the class clusters of test data. These findings suggest that distilled data learns analogous patterns that are present in the real data.

\begin{figure}
    \centering
    \resizebox{0.99\linewidth}{!}{\includegraphics{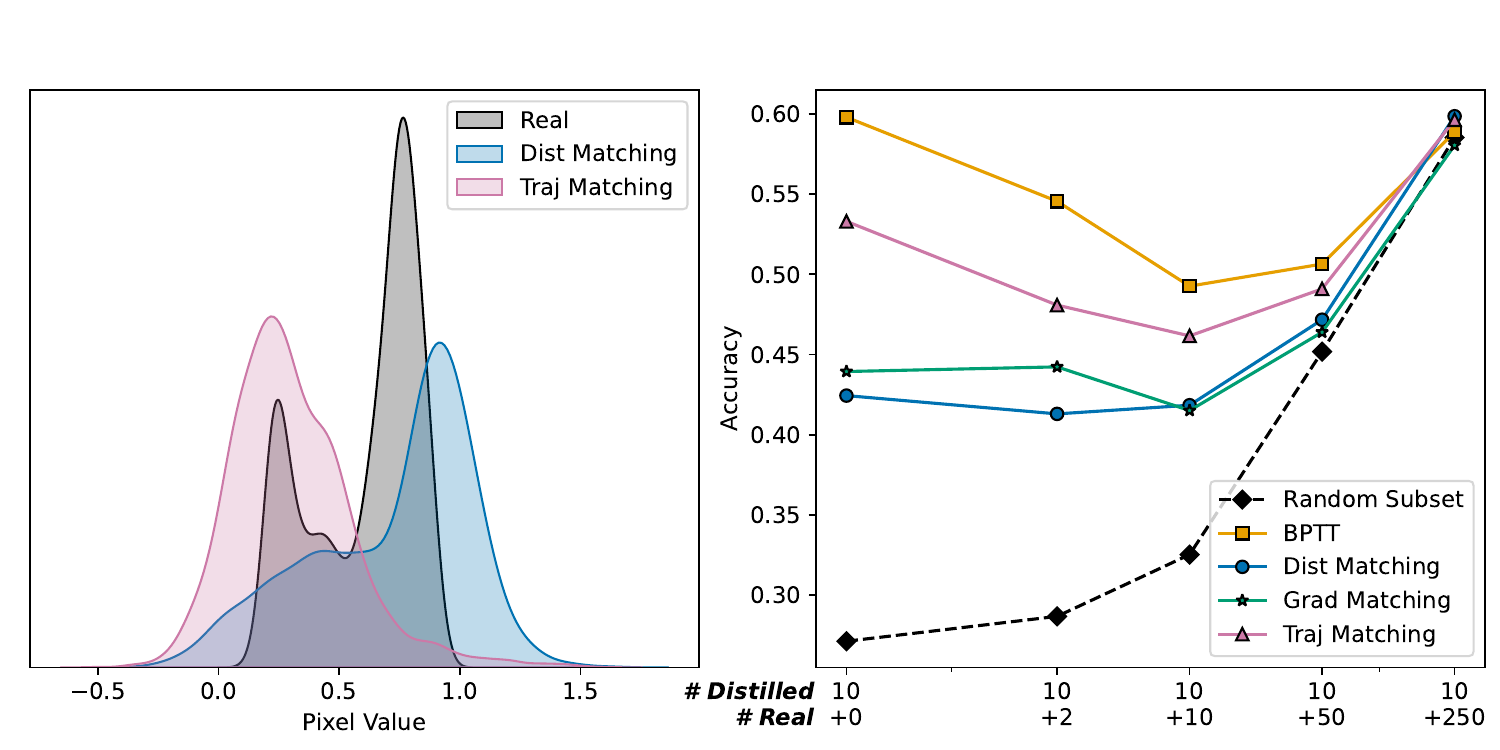}}
    \caption{\textbf{Distilled data is different than real data.} \textit{left.} Kernel density estimation (KDE) plot of pixel intensity of three sample images: a real image, an image distilled with trajectory matching, and an image distilled with distribution matching. Both distilled images contain pixel values outside [0,1]. \textit{right.} Accuracy of models trained on distilled data and real data mixed together. We train models with 10 distilled images (from four different distillation algorithms; different color lines) combined with a random subset of 0-250 real images per class (x-axis). Adding the real data samples into the training does not substantially benefit -- and even in some cases \emph{decreases} -- the accuracy of the trained model! In stark contrast is the baseline (dashed line) trained on 10-260 random real images; it significantly improves with more real data. 
    }
    \label{fig:data}
\end{figure}

\smallsec{Distilled data may not lie on the real data manifold} This is evident by looking at the distribution of pixel intensity between a real image and a distilled image shown in Figure \ref{fig:data} \textit{left}. Real images contain RGB values from 0 to 1. The distilled image shown contains pixel intensity values outside of this range. Additionally, these values are not inconsequential. In the case of BPTT, clipping the pixel intensity outside of the range to between 0 and 1 results in retrained accuracy decrease from 58\% to 44\%.

\smallsec{Distilled data is more sensitive than real data during training} Quantitative differences between distilled images and real images are consequential as they lead to sensitivity of training on distilled data. We demonstrate that when real data is mixed with distilled data during training, the classification performance may \emph{decrease}. In Figure \ref{fig:data} \textit{right}, we train a classifier with both real and distilled data. Adding 2-50 real images to 10 distilled images per class on CIFAR-10 actually \emph{decreases} model accuracy. We show in Section \ref{sec:add_pred_analysis} of appendix that these findings extend to different datasets, data scales, and state-of-the-art (SOTA) dataset distillation algorithms. Thus, training on distilled data is very sensitive and not analogous to training on real data. 

\section{Information Captured in Distilled Data}
\label{sec:info_distilled}
Given the power but also the limitations of distilled data (described in the section above), the question on the exact mechanism of distillation naturally emerges: \textit{What kind of information is captured in distilled data?}

We can reason about distilled data by analogy to how we reason about trained models: distilled data stores enough task-specific information about a training dataset as to be able to generalize to making predictions on a test dataset -- much like the parameters of a trained model do. 
However, having access to distilled data is not sufficient to reach the same generalization accuracy as having access to the parameters of a model trained on the full real dataset. Thus, we can argue that distilled data does not fully capture the task-specific information about the training data distribution, and ask why. One possibility is that distilled data functions similarly to an over-regularized model, e.g., it is storing information similar to what would be learned by an under-powered model. Another possibility is that distilled data is similar to a model trained on only a small subset of training data, e.g., it is simply retaining information about only a subset of the data seen during distillation. 

Through a series of analyses, we demonstrate that \textbf{distilled data appears to capture the same information learned from real data in the early training process of a model}. We first demonstrate the strong parallels between models trained by distilled data and models trained \emph{with early stopping} on real data (which is akin to heavy regularization) by looking at similarity in the model's predictions on the real test data. We provide further evidence of this equivalence by showing successful class recognition on the distilled data is learned early on when training a model on real data. Finally, we deepen this finding by studying the loss curvature with respect to the distilled data on a model trained on real data. We illustrate that this loss curvature with respect to the data distilled by BPTT and trajectory matching quickly becomes flat (low curvature) withing 1-2 epochs of training. The flat region suggests that distilled data captures little information beyond what would already be learned by a model early in the training process. 

\begin{figure}
    \centering
    \includegraphics[width=0.91\linewidth]{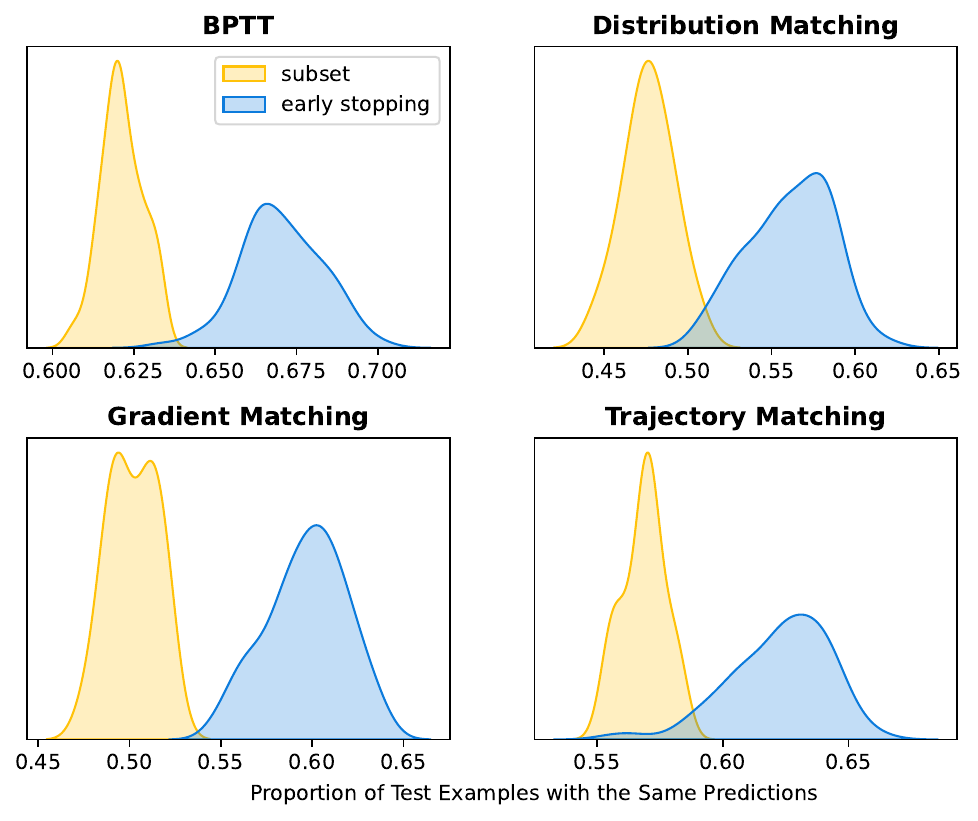}
    \caption{\textbf{Distribution of prediction agreement on CIFAR-10.} Kernel density estimation plots on the number of examples in CIFAR-10 where models that are trained on all of the real data but early stopped or models that are trained on a subset of real data agrees with the model trained on distilled data. The distribution reveals that across all four distillation methods tested, models that are early stopped has a considerable higher number of agreements, indicating that models trained on distilled data predict similarly to models that are early stopped rather than trained on subsets of real data. The similarity with early-stopped models suggests that training on distilled data is analogous to early stopping on real data.}
    \label{fig:agreement}
\end{figure}
\subsection{Predictions of models trained on distilled data is similar to models trained with early-stopping} We first start unraveling the underlying information in distilled data with an analysis on the predictions of models that are trained on different data sources and techniques. Specifically, we analyze predictions of models trained by distilled data (distilled-trained-models) and measure their agreements with predictions from models trained on random subsets of real data (subset-trained-models) and models trained on real data that is early stopped (early-stopped-models). To remove the effect of task accuracy, we only directly compare models with similar test accuracy.  

 For each of the dataset distillation algorithms, we individually compare the distilled-trained-model to the subset-trained-models. We accomplish this by building a pool of 520 models trained on random subsets sampled between 0.5\% to 5\% and training the model for 100 epochs. We only select models with accuracy $\pm 1\%$ compared to the distilled-trained-model, which corresponds to between 20-30 subset-trained-models that are compared with each distilled-train-model. Similarly, we compare the model trained on the distilled data with models trained on whole dataset but are early stopped at an iteration with the closest accuracy (iteration 35 for gradient matching, 40 for distribution matching, 90 for trajectory matching and 130 for BPTT). 
 
The distribution of agreements shown in Figure \ref{fig:agreement} reveals a clear separation in the number of agreements: distilled-trained-models tend to agree more with early-stopped-models rather than subset-trained-models. We provide further support on different dataset, data scale, and SOTA dataset distillation algorithms in section \ref{sec:add_pred_analysis} of the appendix. Additionally, in the same section of the appendix, we repeat the analysis of utilizing models trained on whole dataset but with weight decay regularization to show that early-stopped models is still the most similar to distilled-trained-models.  Finally, we provide additional significance test on the similarity in predictions in section \ref{sec:variability} of the appendix. All of these analyses reveal that distilled-trained-models is analogous to early-stopped-models.  

\begin{figure}
    \centering
    \includegraphics[width=0.95\linewidth]{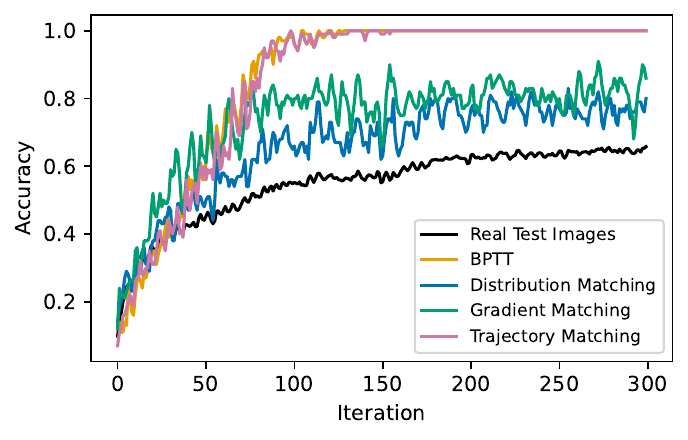}
    \caption{\textbf{Recognition performance on real and distilled data on model trained on real data.} We train a model on real data for 300 iterations and evaluate the model's evaluation accuracy at every iteration of the training. The plot shows classification accuracy on BPTT, distribution matching, gradient matching, and trajectory matching distilled data stops improving after iteration 150 but the classification accuracy on real test still improve. The lack of improvement of classification accuracy on distilled data shows the information that the model learns relevant to correctly classifying the distilled data exists only in the early iterations of training on real data. Therefore, this suggests that distilled data stores information regarding the early training dynamics of real data.}
    \label{fig:accuracy_over_time}
\end{figure}

\subsection{Recognition on the distilled data is learned early in the training process} So far we were able to claim the predictions of models trained on distilled data correlate with the predictions of models trained on real data but with early stopping. However, this does not necessarily prove that the models are capturing the same information. To dig deeper, we pose the following question: when does a model trained on real data successfully recognize the classes in distilled data? The answer to this question allows us to connect information in distilled data to information a model learns from real data at different stages of training.

To answer the question, we revisit the inverse pipeline: training on real images and evaluating on distilled images. We evaluate the accuracy on classifying the data on the real test data as well as data distilled by BPTT, distribution matching, gradient matching, and trajectory matching. The result in Figure \ref{fig:accuracy_over_time} reveals that performance on distilled data of a model trained on real data stops improving after iteration 150 even though the performance on the real test data is still improving. More specifically, we observe that, by iteration 150, BPTT/trajectory matching distilled data reaches 100\% accuracy and distribution/gradient matching distilled data fluctuate at around 70\%-80\% accuracy while the accuracy on the real test images is still increasing. This indicates that the information learned from the real training data after iteration 150 is not relevant to the distilled data. Hence, the stagnant recognition performance on distilled data in later training suggests information captured by distilled data is only relevant to the early training process of real data.

\begin{figure*}[t]

\centering
\includegraphics[width=0.9\linewidth]{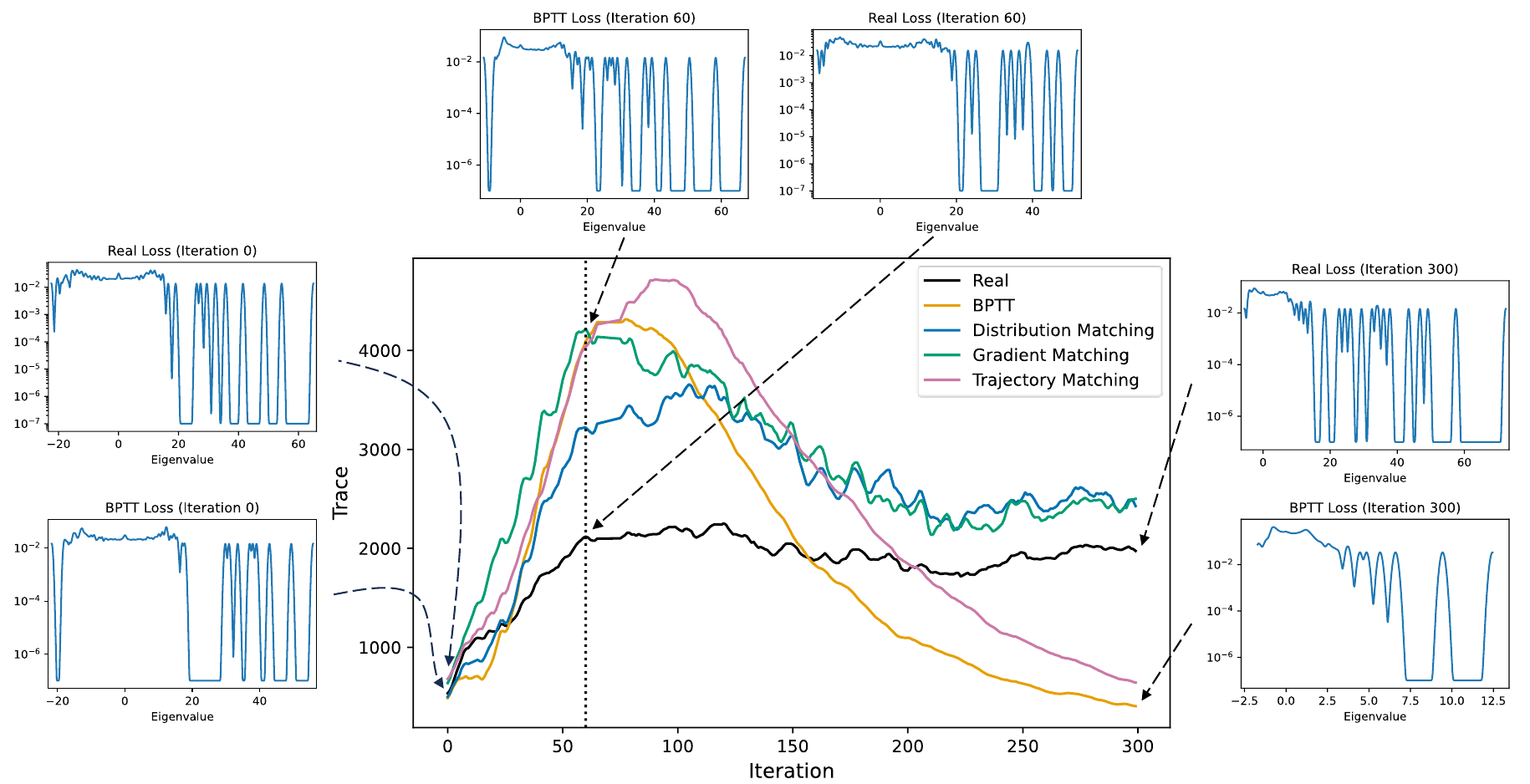}
\caption{\textbf{Curvature of loss landscapes with real-vs-distilled  data.} We show that a model trained on \emph{real} data quickly learns the information contained in distilled images (here in 300 iterations, about 1.5 epochs). For each training iteration we evaluate the loss of the model with respect to five types of data: its training set of real images and four sets of distilled images (distilled from the same training set). For each, we compute the loss curvature and report the smoothed trace of the corresponding Hessian matrix using Hutchinson's method~\cite{hutchinson1989stochastic} implemented in PyHessian~\cite{yao2020pyHessian}. This summarizes the local curvature:
high trace values correspond to regions of high loss curvature (rapidly changing gradient with respect to the data, typically seen during iterations of learning) and low trace values correspond to either flat regions (typically seen at convergence) or saddle point regions (typically seen at the beginning of the training, with a high-curvature landscape but in different  directions). 
To differentiate between saddle point and flat regions, the side plots show the log density of the eigenvalues of the Hessian  with respect to the real train data and BPTT distilled data using the stochastic Lanczos quadrature algorithm \cite{golub1969calculation, ghorbani2019investigation}. From iteration 0-50, the loss curvature indicates a saddle point (mix of positive and negative eigenvalues). In iteration 50-150, the curvature induced by the distilled data becomes progressively higher than the curvature induced by the real train data, indicating that the model (while being trained on real images) is rapidly learning the information captured in the distilled data. Finally, the curvature induced by BPTT/trajectory matching distilled data reaches a flat region (with low-magnitude eigenvalues, as seen in the side plot for BPTT). This suggests that a model trained on real data for 300 iterations wouldn't learn much new information if its training is continued using BPTT distilled data. The peak in curvature followed by the flat region within 300 iterations suggest that the BPTT and trajectory matching distilled data only capture the information about the training data that would be learned early in the training process. Unfortunately we can't draw similar conclusions for distribution/gradient matching data, as the trace fluctuates greatly between iterations. 
}
\label{fig:eigen}
\end{figure*}

\subsection{Distilled data stores little information beyond what would be learned early in training} Recognition accuracy alone is not sufficient to fully uncover the underlying information as behavior of distilled data can be different between training and inference as shown in Section \ref{sec:distilledvsreal}. Therefore, we dig deeper by comparing the curvature induced by distilled data vs. real data. To accomplish this, we inspect the Hessian Matrix, which describes the local curvature. Analysis with the Hessian Matrix is also advantageous because it closely related to Fisher Information, which is used in previous works to study the information learned by a model during training \cite{achille2018critical}.

Figure \ref{fig:eigen} reveals that the curvature of loss induced by BPTT and trajectory matching distilled data can be categorized into three stages: initial training starts on a saddle point, followed by region of high curvature, and ending at a low curvature flat region. The initialization on a saddle point is expected and well aligned with the curvature of the real data. The region of high curvature suggests that data distilled by BPTT and trajectory matching are highly informative to the model during those iterations, i.e. large changes in the model parameters if trained on such distilled data during these stages, More importantly, the flat region in the later iterations indicates little information provided by the distilled data. In other words, the model that is trained on real data after certain number of iterations would not change significantly if it was trained additionally on data distilled by BPTT and trajectory matching. Therefore, the flat regions reveal that BPTT and trajectory matching distilled data contain only information of the early training dynamics. We provide further support and intuition through loss landscape visualization and additional curvature analysis on data distilled with different storage budget in section \ref{sec:add_loss_analysis} of the appendix. We also empirically verify our intuition on the connection between flat regions and task performance from additional training  as well as an explanation on the high curvature induced by distribution matching and gradient matching distilled data in section \ref{sec:additional_training} of the appendix.

\section{Semantics of Captured Information}
\label{sec:influence}
\begin{figure}[t]
    \centering
    \includegraphics[width=\linewidth]{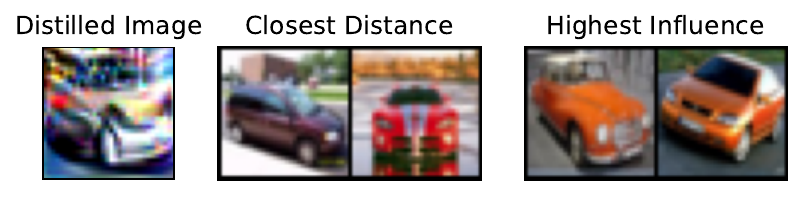}
    \includegraphics[width=0.99\linewidth]{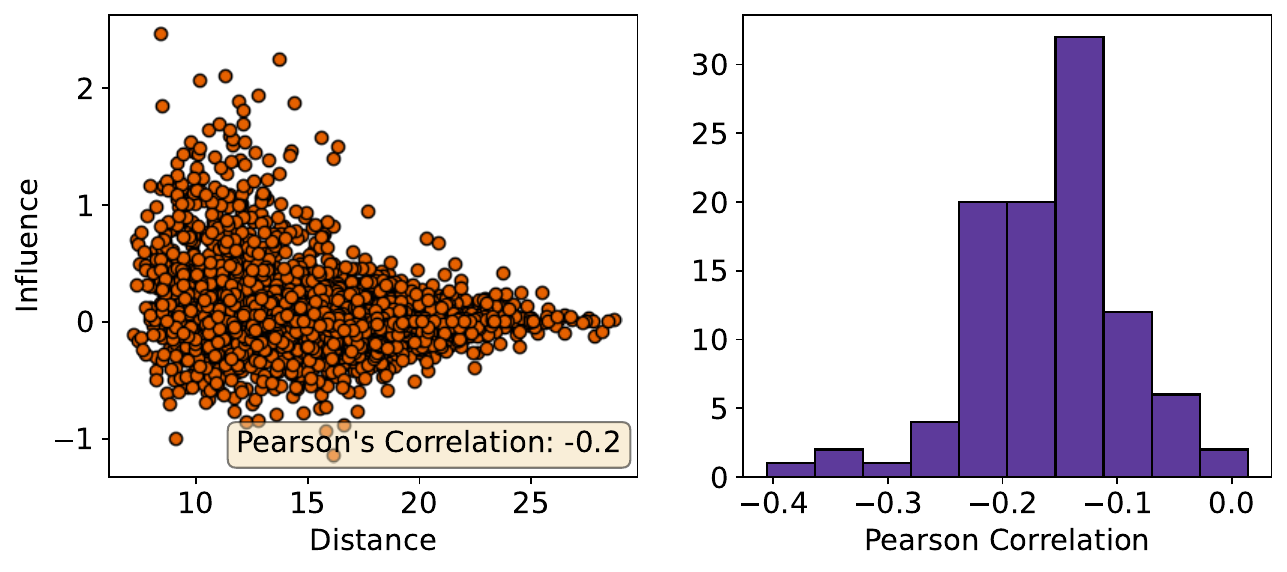}
    \caption{\textbf{Influence is more than visual similarity.} \textit{top.} An example image distilled with trajectory matching, two of its closest test images using the penultimate features of a trained ResNet-18 model and two test images with highest influence. \textit{bottom left.} Using the same distilled image as \textit{top}, we visualize the feature distance vs. the influence and see that there is  very weak correlation between the two (although very high-influence  images do tend to be more visually similar). \textit{bottom right.} Pearson correlation between feature distance vs. influence on each of the 100 distilled images compared to all the real CIFAR-10 test images, confirming the observation in \emph{bottom left}. Hence, visual similarity cannot fully explain the influence observed by distilled images.}
    \label{fig:similarity}
\end{figure}
Since distilled data is compressing the learning trajectories, \textit{do distilled data points individually carry meaningful information?} Visualizing the distilled image provides little insight on the information contained in it (see Figure~\ref{fig:hook}). Thus, to answer this question, we build an interpretable framework by leveraging ideas from influence function to isolate information compressed in each individual distilled data point.  
We reveal that each distilled data point does contain meaningful semantic information. 

\begin{figure}
    \centering
    \includegraphics[width=0.99\linewidth]{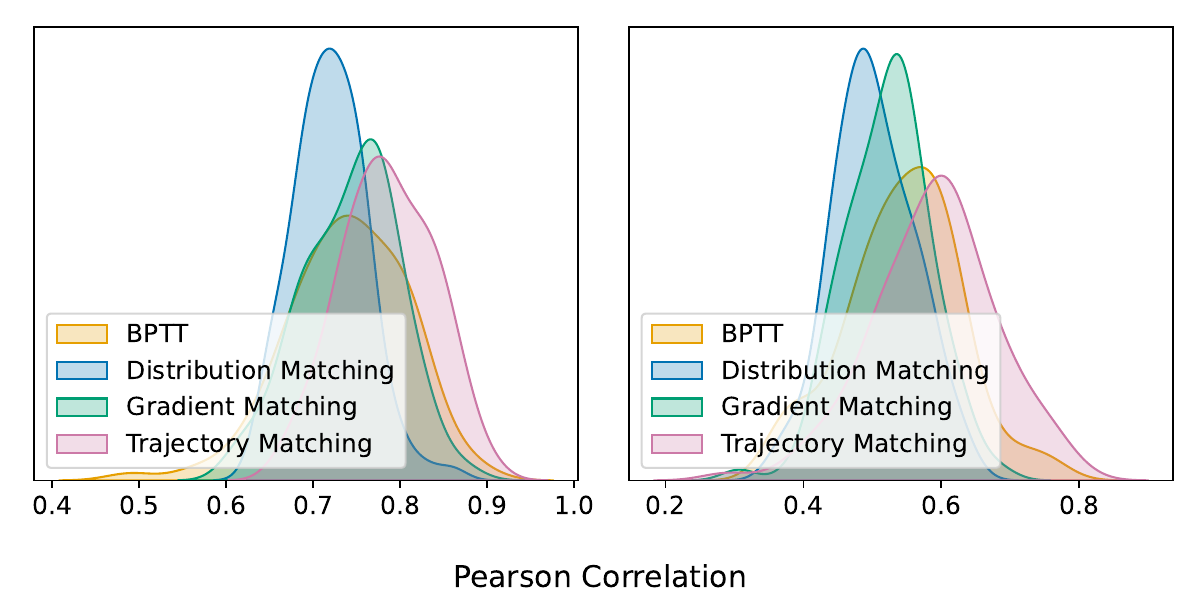}
    \caption{\textbf{Influence of distilled images is consistent.} \textit{left.} Kernel density estimation plot of distribution of Pearson correlation between influences calculated across two different random seeds. The high correlation confirms that the observed influence  is due to distilled image rather than random chance. \textit{right.} The correlation between the influence of a distlled image calculated with leave-one-out retraining when using 100 real images versus when using the 99 other distilled images. The moderately high correlations suggest that the influence of a distilled image extends to real data training, and hence, the information compressed is universal and not contingent on other distilled images.}
    \label{fig:consistency}
\end{figure}

\subsection{Influence functions for understanding distilled data}
The influence function is an concept proposed to quantify the impact of individual data points on a model's predictions \cite{koh2017understanding}. This measure is often used to better understand the model's decision process and can identify potential errors in a model's prediction \cite{wang2023error}. In contrast, we leverage influence function to better understand the nature of images generated using dataset distillation. 

\smallsec{Exact computation} The influence function in machine learning was originally proposed with first-order approximations using implicit Hessian-vector products given the large computation cost associated with retraining models on large datasets \cite{koh2017understanding}. The accuracy of the approximation was originally verified through comparison with leave-one-out retraining, but subsequent works have shown that such approximation can be fragile in different settings \cite{basu2020influence}. Since dataset distillation operates in a low-data setting, we circumvent the approximation entirely and directly calculate influence with leave-one-out retraining. Formally, we calculate the influence of distilled image $x_d$ on test image $x_t$ where $\hat\theta$ is the model learned using all the distilled images and $\hat\theta_{-x_d}$ is model learned with all but the $x_d$ distilled image with loss function $L$ as 
$$ I_{x_d \rightarrow x_t} = L(x_t; \hat\theta_{-x_d}) - L(x_t; \hat\theta). $$

\smallsec{Influence is not image similarity} To confirm the necessity of calculating the influence for examination of information content, in Figure~\ref{fig:similarity}, we verify that influence functions reveal information that cannot be glimpsed through visual similarity alone. Although real images $x_t$ which are identified as being highly influential do tend to be somewhat visually similar to the corresponding distilled image $x_d$ (Figure~\ref{fig:similarity} \emph{bottom right}), overall there is only a very weak correlation between the two measures. 

Looking at a visual example of a distilled image along with two highly-ranked real images (by closest distance and highest influence; Figure~\ref{fig:similarity} \emph{top}), we begin to notice that highest influence images do appear to be semantically consistent (e.g., both are images of \emph{orange} cars). In Sections~\ref{sec:dd_infodecomposition} we leverage this insight to begin to understand the  information content of distilled images.

\smallsec{Influence is consistent across random training runs} We further perform a simple sanity check to ensure the observed influence is consistent and does not change across multiple runs. To check this, we evaluated the influence of the distilled image trained using a different random seed and measure the Pearson correlation against the original run. The result shown in Figure \ref{fig:consistency} \textit{left} reveals that the influence between different runs are highly correlated. Hence, the highly influenced images observed are caused by the distilled image rather than by chance. 

\smallsec{Influence computed for each distilled image is independent of other distilled images} Finally, we check that the computed influence  generalizes and is not dependent on other distilled images. To test this, for each distilled image,
we utilize 100 additional real images (instead of the 99 other distilled images) to perform leave-one-out retraining. We re-calculate influence as the difference in loss with the model trained on the 100 real data points and the model trained on the 100 real data points and the distilled data point. The correlation revealed in Figure~\ref{fig:consistency} \emph{right} suggests the calculated influence is quite consistent even in this new setting, and not (too) dependent on other distilled data points.

\begin{figure}[t]
    \centering
    \includegraphics[width=0.99\linewidth]{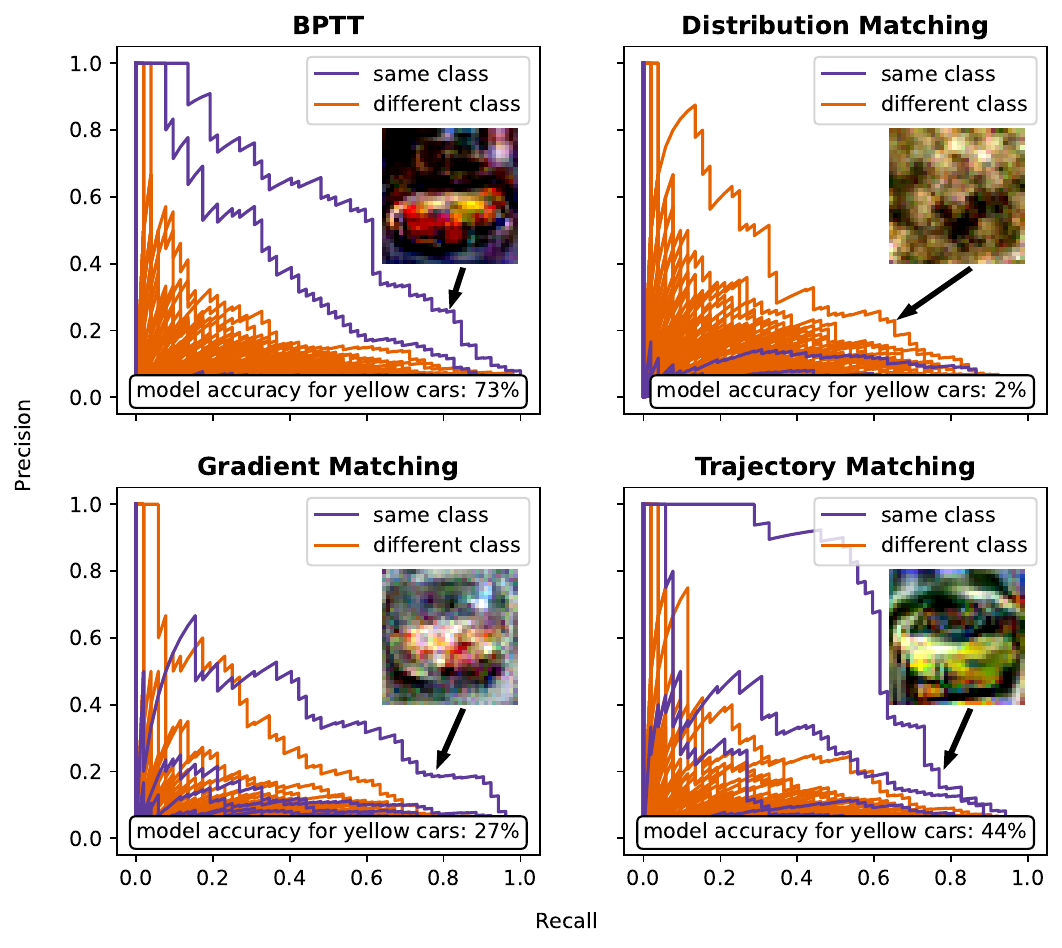}
    \caption{\textbf{Precision Recall curves on yellow cars.} PR curves of different distilled images $x_d$, evaluating the computed influence function $I_{x_d\rightarrow x_t}$ at classifying test images of cars $x_t$ as belonging to a \emph{yellow} car. The distilled image $x_d$ that results in the highest area under the curve is shown on the plot.  The curves reveal that (particularly for BPTT and trajectory matching) some distilled images do capture the concept of yellow cars. At the bottom of each plot, we further report the accuracy of a distilled-data-trained model on classifying real-world images of yellow cars as being a ``car''. This accuracy is somewhat higher for distillation methods where the ``yellow car'' concept appears to be captured in individual distilled examples, providing further support for the findings. }
    \label{fig:precisionrecall}
\end{figure}

\subsection{Distilled data contains semantic information beyond the class label} 
\label{sec:dd_infodecomposition}
For a more comprehensive analysis, we directly calculate the information stored in the distilled data and find the influence of specific distilled data points is predictive of non-class related semantics attributes, indicating that  semantic concepts (beyond the class label) are stored in individual distilled data points.

\smallsec{Semantic extraction} CIFAR-10 is a rather simple dataset that lacks any additional metadata information of its images outside of the class labels. To overcome this limitation, we leverage current breakthroughs in large foundation models to assign additional semantic information to each of the images. We use LLaVA \cite{liu2023improvedllava,liu2023llava}, a state-of-the-art multimodal model, for the assignment. We first query each image with the prompt ``describe the object" and ``describe the background" to generate a pool of candidate semantic attributes. We extract the corresponding attributes by manually inspecting the response from the highly influenced real images to compile a list of reoccurring attributes.  

For each distilled image, its influence score can be used to rank the real test images, where each real image has been  annotated with semantic attributes as above. The area under the precision-recall curve for a particular semantic attribute can then be used to determine if this distilled image strongly influences real images which contain this  attribute. 

\begin{table}[t]

\centering
\caption{\textbf{BPTT distilled image semantics.} Examples of the extracted semantics of nine notable images distilled with BPTT.}
\resizebox{0.89\linewidth}{!}{%
\begin{tabular}{c|c|c}
\hline
Image & Class & Semantics                          \\ \hline
\begin{tabular}{@{}c@{}}\includegraphics[width=0.2\linewidth]{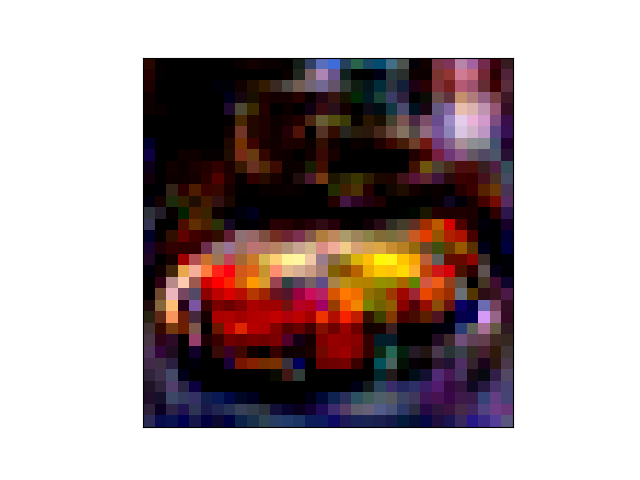}\end{tabular} & Car   & Yellow                             \\
\begin{tabular}{@{}c@{}}\includegraphics[width=0.2\linewidth]{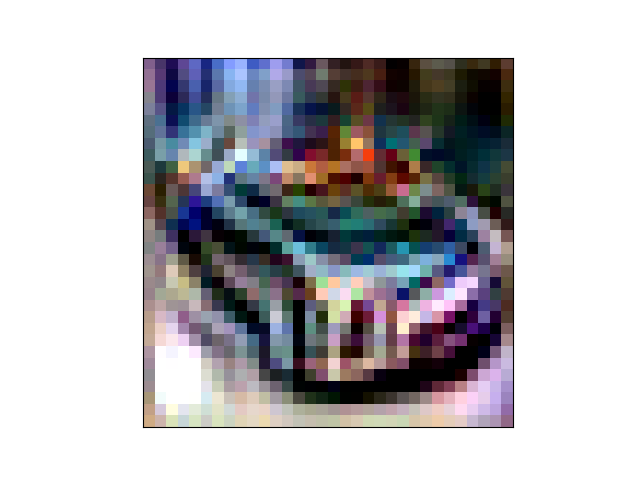}\end{tabular}                & Car   & Parking Lot, Garage                \\
\begin{tabular}{@{}c@{}}\includegraphics[width=0.2\linewidth]{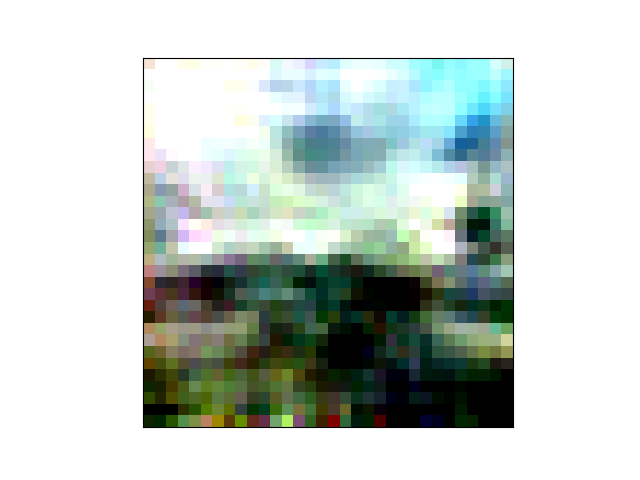}\end{tabular}                & Plane & Parked, Ground, Runway             \\
\begin{tabular}{@{}c@{}}\includegraphics[width=0.2\linewidth]{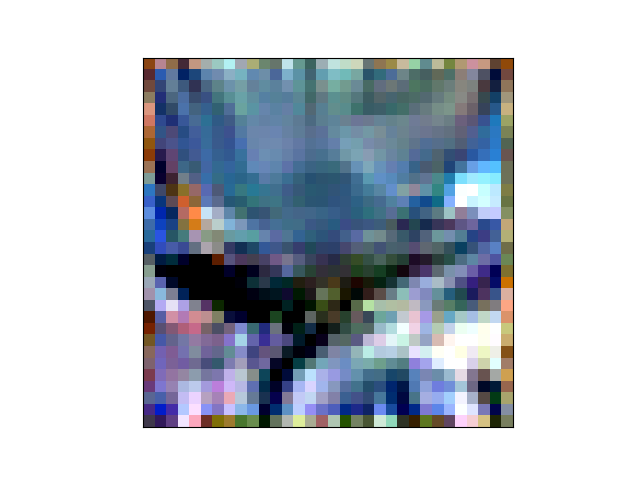}\end{tabular}                & Plane & Jet flying through the sky         \\
\begin{tabular}{@{}c@{}}\includegraphics[width=0.2\linewidth]{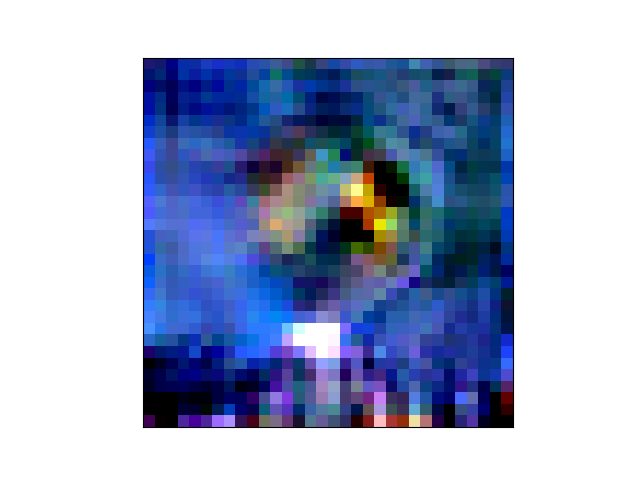}\end{tabular}               & Bird  & Flying through the sky, blue sky \\ 
\begin{tabular}{@{}c@{}}\includegraphics[width=0.2\linewidth]{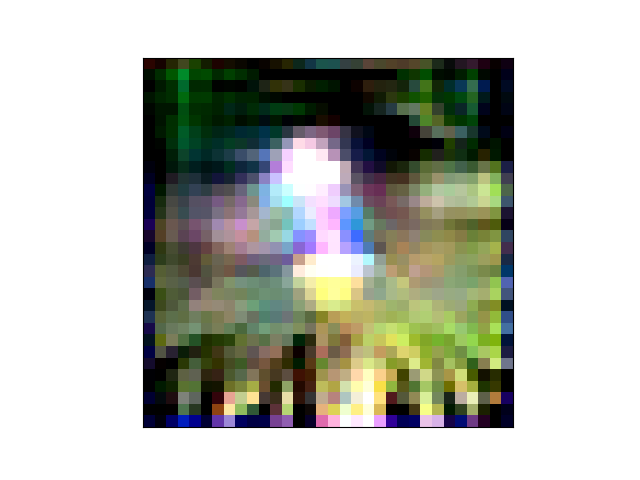}\end{tabular} & Bird  & Grassy field, tree, green          \\
\begin{tabular}{@{}c@{}}\includegraphics[width=0.2\linewidth]{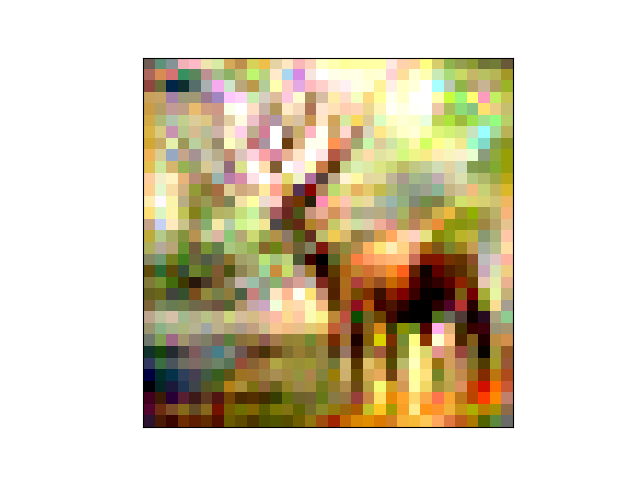}\end{tabular}               & Deer  & Grassy/snowy field                 \\
\begin{tabular}{@{}c@{}}\includegraphics[width=0.2\linewidth]{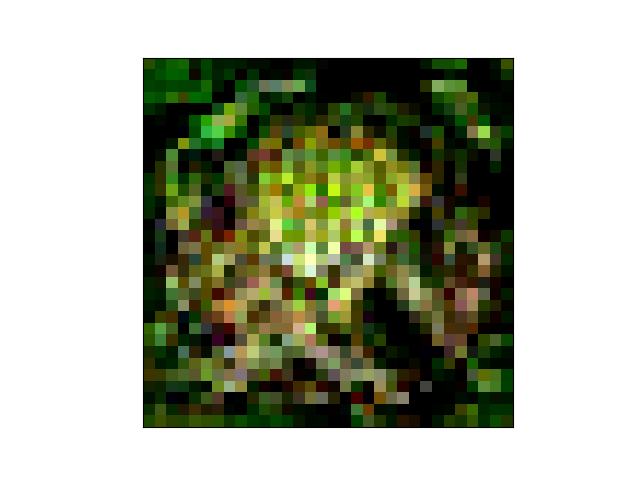}\end{tabular}               & Frog  & Muddy, rocky, dark background     \\
\begin{tabular}{@{}c@{}}\includegraphics[width=0.2\linewidth]{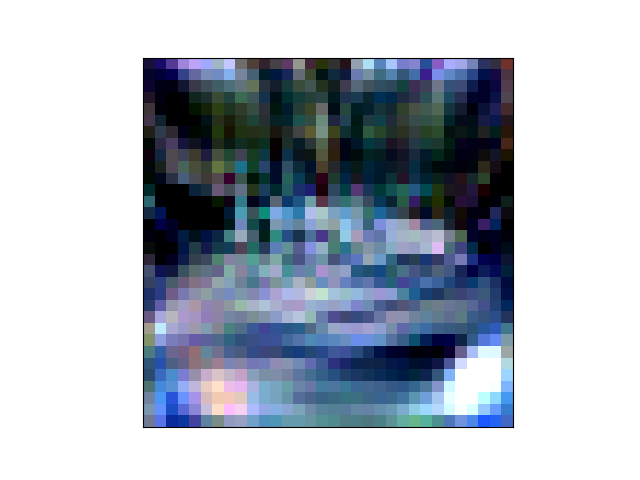}\end{tabular}               & Boat  & Forest, lake, parked               \\ \hline
\end{tabular}%
}
\label{table:examples}
\end{table}

\smallsec{Quantitative analysis} We utilize the precision-recall curve to identify distilled images that contain information related to specific semantics. We illustrate an example of the precision recall curve on the semantic \textit{yellow} in images of cars in Figure \ref{fig:precisionrecall}. The precision recall curve reveals that for some distilled images, the information compressed is heavily associated to the semantics of yellow cars. In particular, we observe two distilled image with notably higher precision recall curves from BPTT and one distilled image from trajectory matching. Interestingly, the existence of these highly predicative distilled image also corresponds with the classification performance of models trained, where models trained BPTT have high accuracy on yellow cars while models trained on distribution matching has low accuracy. We extend this analysis across all 100 distilled images from BPTT and show that semantics can be extracted from other distilled images in Table \ref{table:examples}.

\begin{figure*}
    \centering
    \includegraphics[width=0.99\linewidth]{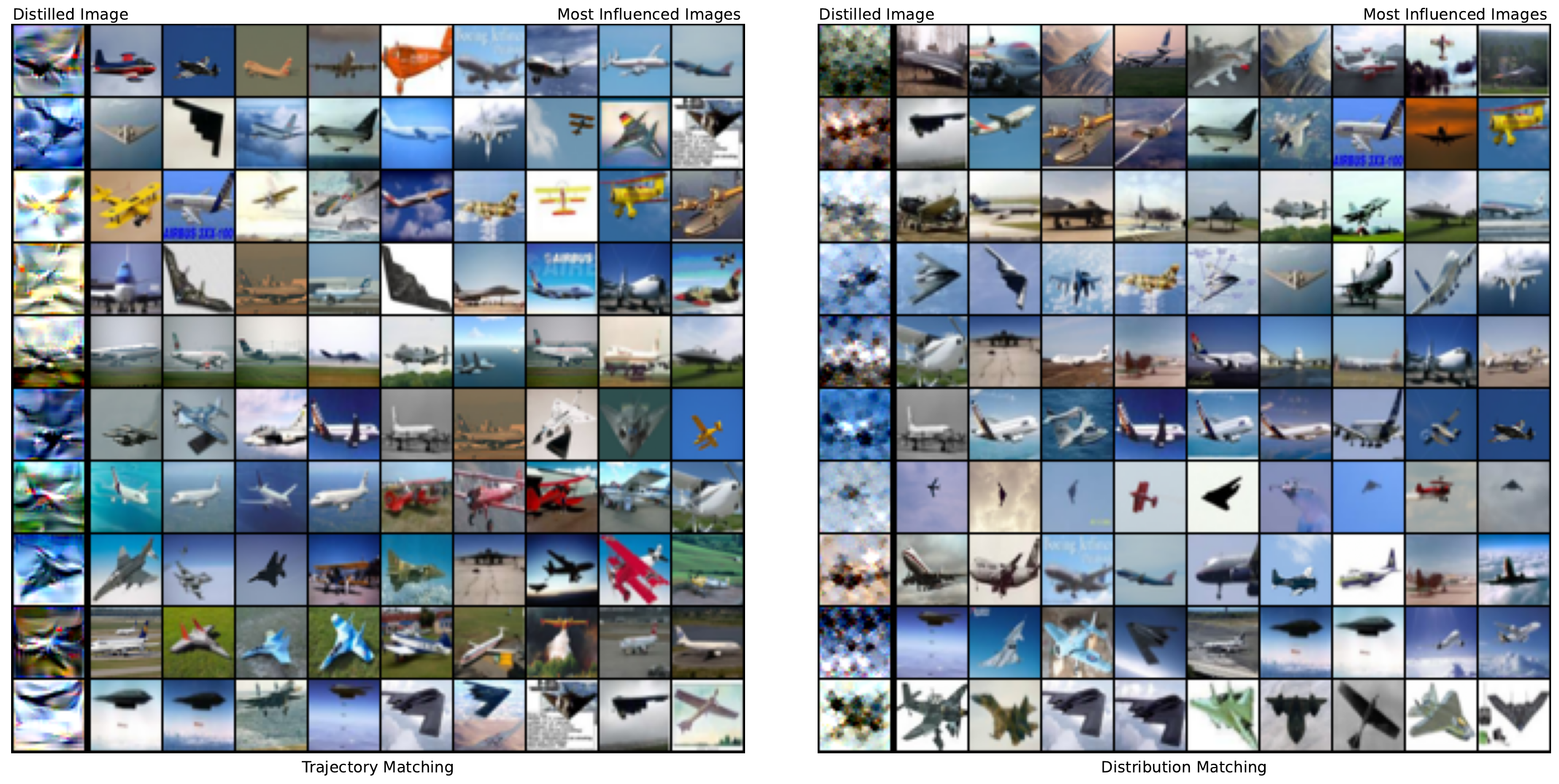}
    \caption{\textbf{Distilled images of airplanes and the highly influenced test images.} The plot illustrates the nine most influenced images in CIFAR-10 test by images distilled with trajectory matching and distribution matching. The images influenced are heterogeneous across the different distilled images while homogeneous within the same distilled image. In consequence, while distilled data themselves are uninterpretable, the images influenced are, and hence, can be used as a method for interpreting information captured. }
    \label{fig:airplane}
\end{figure*} 

\smallsec{Qualitative analysis}
Finally, we perform qualitative analysis and display 10 images distilled using trajectory matching and distribution matching and nine corresponding images from CIFAR-10 test that are most influenced in Figure \ref{fig:airplane}. First, visualization reveals the influenced images across different distilled images are very heterogeneous. For example, in trajectory matching, the last distilled image appears to be related to stealth military planes in the sky while the second to last distilled image appears to be related to planes landed on the ground. A similar finding can be seen with distribution matching where the last two distilled images represent stealth military planes but display a blue sky in the background and a white background respectively. 

Additionally, we observe homogeneity in the images influenced from some of the distilled images. For example, the real images influenced by the last distilled image from both trajectory and distribution matching mostly consist of military style aircraft. Multiple homogeneous groups can also be associated with each distilled image. For example, the sixth distilled image from trajectory matching presents airliner flying across the ocean but also red biplane. The heterogeneity in influenced images across different distilled images with the homogeneity in influenced images from the same distilled image suggests that different semantic concepts are in different distilled images. 

\section{Conclusion}
  Through our investigation, we have addressed three key questions. First, we showed that while distilled data behave like real data at inference time, they are sensitive to the training procedure and cannot serve as drop in replacements for real data. Second, we demonstrated that dataset distillation captures the early learning dynamics of real models. Finally, we revealed that individual distilled data points encapsulate meaningful semantic information. Our study offers valuable insights into underlying mechanisms of dataset distillation, informing better design of future methods.

\section*{Impact Statement}
 Dataset distillation can enable compression of large-scale data into smaller, more resource efficient, and thus more democratized datasets. However, there are important concerns regarding whether and how dataset biases get captured, amplified, and/or propagated within distilled datasets. Although we do not tackle this problem directly, we hope our work can help meaningfully shed insights into such questions.

\section*{Acknowledgements}
This material is based upon work supported by the National Science Foundation under Grant No. 2112562. Any opinions, findings, and conclusions or recommendations expressed in this material are those of the author(s) and do not necessarily reflect the views of the National Science Foundation. We thank Allison Chen, Amaya Dharmasiri, Eashan Garg, Erich Liang, Esin Tureci, Tyler Zhu, and Xindi Wu for proofreading and providing valuable feedback on the manuscript. 

\bibliography{paper}

\begin{thebibliography}{46}
\providecommand{\natexlab}[1]{#1}
\providecommand{\url}[1]{\texttt{#1}}
\expandafter\ifx\csname urlstyle\endcsname\relax
  \providecommand{\doi}[1]{doi: #1}\else
  \providecommand{\doi}{doi: \begingroup \urlstyle{rm}\Url}\fi

\bibitem[Achille et~al.(2018)Achille, Rovere, and Soatto]{achille2018critical}
Achille, A., Rovere, M., and Soatto, S.
\newblock Critical learning periods in deep networks.
\newblock In \emph{International Conference on Learning Representations}, 2018.

\bibitem[Basu et~al.(2020)Basu, Pope, and Feizi]{basu2020influence}
Basu, S., Pope, P., and Feizi, S.
\newblock Influence functions in deep learning are fragile.
\newblock In \emph{International Conference on Learning Representations}, 2020.

\bibitem[Cazenavette et~al.(2022)Cazenavette, Wang, Torralba, Efros, and Zhu]{cazenavette2022dataset}
Cazenavette, G., Wang, T., Torralba, A., Efros, A.~A., and Zhu, J.-Y.
\newblock Dataset distillation by matching training trajectories.
\newblock In \emph{Proceedings of the IEEE/CVF Conference on Computer Vision and Pattern Recognition}, pp.\  4750--4759, 2022.

\bibitem[Cui et~al.(2023)Cui, Wang, Si, and Hsieh]{cui2023scaling}
Cui, J., Wang, R., Si, S., and Hsieh, C.-J.
\newblock Scaling up dataset distillation to imagenet-1k with constant memory.
\newblock In \emph{International Conference on Machine Learning}, pp.\  6565--6590. PMLR, 2023.

\bibitem[Deng \& Russakovsky(2022)Deng and Russakovsky]{deng2022remember}
Deng, Z. and Russakovsky, O.
\newblock Remember the past: Distilling datasets into addressable memories for neural networks.
\newblock \emph{Advances in Neural Information Processing Systems}, 35:\penalty0 34391--34404, 2022.

\bibitem[Dosovitskiy et~al.(2020)Dosovitskiy, Beyer, Kolesnikov, Weissenborn, Zhai, Unterthiner, Dehghani, Minderer, Heigold, Gelly, et~al.]{dosovitskiy2020image}
Dosovitskiy, A., Beyer, L., Kolesnikov, A., Weissenborn, D., Zhai, X., Unterthiner, T., Dehghani, M., Minderer, M., Heigold, G., Gelly, S., et~al.
\newblock An image is worth 16x16 words: Transformers for image recognition at scale.
\newblock \emph{arXiv preprint arXiv:2010.11929}, 2020.

\bibitem[Ghorbani et~al.(2019)Ghorbani, Krishnan, and Xiao]{ghorbani2019investigation}
Ghorbani, B., Krishnan, S., and Xiao, Y.
\newblock An investigation into neural net optimization via hessian eigenvalue density.
\newblock In \emph{International Conference on Machine Learning}, pp.\  2232--2241. PMLR, 2019.

\bibitem[Glorot \& Bengio(2010)Glorot and Bengio]{glorot2010understanding}
Glorot, X. and Bengio, Y.
\newblock Understanding the difficulty of training deep feedforward neural networks.
\newblock In \emph{Proceedings of the thirteenth international conference on artificial intelligence and statistics}, pp.\  249--256. JMLR Workshop and Conference Proceedings, 2010.

\bibitem[Golub \& Welsch(1969)Golub and Welsch]{golub1969calculation}
Golub, G.~H. and Welsch, J.~H.
\newblock Calculation of gauss quadrature rules.
\newblock \emph{Mathematics of computation}, 23\penalty0 (106):\penalty0 221--230, 1969.

\bibitem[Goodfellow et~al.(2014)Goodfellow, Vinyals, and Saxe]{goodfellow2014qualitatively}
Goodfellow, I.~J., Vinyals, O., and Saxe, A.~M.
\newblock Qualitatively characterizing neural network optimization problems.
\newblock \emph{arXiv preprint arXiv:1412.6544}, 2014.

\bibitem[Gretton et~al.(2012)Gretton, Borgwardt, Rasch, Sch{\"o}lkopf, and Smola]{gretton2012kernel}
Gretton, A., Borgwardt, K.~M., Rasch, M.~J., Sch{\"o}lkopf, B., and Smola, A.
\newblock A kernel two-sample test.
\newblock \emph{The Journal of Machine Learning Research}, 13\penalty0 (1):\penalty0 723--773, 2012.

\bibitem[Guo et~al.(2022)Guo, Zhao, and Bai]{guo2022deepcore}
Guo, C., Zhao, B., and Bai, Y.
\newblock Deepcore: A comprehensive library for coreset selection in deep learning.
\newblock In \emph{International Conference on Database and Expert Systems Applications}, pp.\  181--195. Springer, 2022.

\bibitem[He et~al.(2016)He, Zhang, Ren, and Sun]{he2016deep}
He, K., Zhang, X., Ren, S., and Sun, J.
\newblock Deep residual learning for image recognition.
\newblock In \emph{Proceedings of the IEEE conference on computer vision and pattern recognition}, pp.\  770--778, 2016.

\bibitem[Hutchinson(1989)]{hutchinson1989stochastic}
Hutchinson, M.~F.
\newblock A stochastic estimator of the trace of the influence matrix for laplacian smoothing splines.
\newblock \emph{Communications in Statistics-Simulation and Computation}, 18\penalty0 (3):\penalty0 1059--1076, 1989.

\bibitem[Jiang et~al.(2023)Jiang, Gu, Liu, and Pan]{jiang2023delving}
Jiang, Z., Gu, J., Liu, M., and Pan, D.~Z.
\newblock Delving into effective gradient matching for dataset condensation.
\newblock In \emph{2023 IEEE International Conference on Omni-layer Intelligent Systems (COINS)}, pp.\  1--6. IEEE, 2023.

\bibitem[Kim et~al.(2022)Kim, Kim, Oh, Yun, Song, Jeong, Ha, and Song]{kim2022dataset}
Kim, J.-H., Kim, J., Oh, S.~J., Yun, S., Song, H., Jeong, J., Ha, J.-W., and Song, H.~O.
\newblock Dataset condensation via efficient synthetic-data parameterization.
\newblock In \emph{International Conference on Machine Learning}, pp.\  11102--11118. PMLR, 2022.

\bibitem[Kingma \& Ba(2014)Kingma and Ba]{kingma2014adam}
Kingma, D.~P. and Ba, J.
\newblock Adam: A method for stochastic optimization.
\newblock In \emph{International Conference on Learning Representations}, 2014.

\bibitem[Koh \& Liang(2017)Koh and Liang]{koh2017understanding}
Koh, P.~W. and Liang, P.
\newblock Understanding black-box predictions via influence functions.
\newblock In \emph{International conference on machine learning}, pp.\  1885--1894. PMLR, 2017.

\bibitem[Krizhevsky et~al.(2009)Krizhevsky, Hinton, et~al.]{krizhevsky2009learning}
Krizhevsky, A., Hinton, G., et~al.
\newblock Learning multiple layers of features from tiny images.
\newblock 2009.

\bibitem[Krizhevsky et~al.(2012)Krizhevsky, Sutskever, and Hinton]{krizhevsky2012imagenet}
Krizhevsky, A., Sutskever, I., and Hinton, G.~E.
\newblock Imagenet classification with deep convolutional neural networks.
\newblock \emph{Advances in neural information processing systems}, 25, 2012.

\bibitem[Le \& Yang(2015)Le and Yang]{le2015tiny}
Le, Y. and Yang, X.
\newblock Tiny imagenet visual recognition challenge.
\newblock \emph{CS 231N}, 7\penalty0 (7):\penalty0 3, 2015.

\bibitem[Li et~al.(2018)Li, Xu, Taylor, Studer, and Goldstein]{li2018visualizing}
Li, H., Xu, Z., Taylor, G., Studer, C., and Goldstein, T.
\newblock Visualizing the loss landscape of neural nets.
\newblock \emph{Advances in neural information processing systems}, 31, 2018.

\bibitem[Lilliefors(1967)]{lilliefors1967kolmogorov}
Lilliefors, H.~W.
\newblock On the kolmogorov-smirnov test for normality with mean and variance unknown.
\newblock \emph{Journal of the American statistical Association}, 62\penalty0 (318):\penalty0 399--402, 1967.

\bibitem[Liu et~al.(2023{\natexlab{a}})Liu, Li, Li, and Lee]{liu2023improvedllava}
Liu, H., Li, C., Li, Y., and Lee, Y.~J.
\newblock Improved baselines with visual instruction tuning, 2023{\natexlab{a}}.

\bibitem[Liu et~al.(2023{\natexlab{b}})Liu, Li, Wu, and Lee]{liu2023llava}
Liu, H., Li, C., Wu, Q., and Lee, Y.~J.
\newblock Visual instruction tuning.
\newblock In \emph{NeurIPS}, 2023{\natexlab{b}}.

\bibitem[Loshchilov \& Hutter(2018)Loshchilov and Hutter]{loshchilov2018decoupled}
Loshchilov, I. and Hutter, F.
\newblock Decoupled weight decay regularization.
\newblock In \emph{International Conference on Learning Representations}, 2018.

\bibitem[Maalouf et~al.(2023)Maalouf, Tukan, Loo, Hasani, Lechner, and Rus]{maalouf2023size}
Maalouf, A., Tukan, M., Loo, N., Hasani, R., Lechner, M., and Rus, D.
\newblock On the size and approximation error of distilled sets.
\newblock \emph{arXiv preprint arXiv:2305.14113}, 2023.

\bibitem[McInnes et~al.(2018)McInnes, Healy, Saul, and Gro{\ss}berger]{mcinnes2018umap}
McInnes, L., Healy, J., Saul, N., and Gro{\ss}berger, L.
\newblock Umap: Uniform manifold approximation and projection.
\newblock \emph{Journal of Open Source Software}, 3\penalty0 (29):\penalty0 861, 2018.

\bibitem[Nguyen et~al.(2020)Nguyen, Chen, and Lee]{nguyen2020dataset}
Nguyen, T., Chen, Z., and Lee, J.
\newblock Dataset meta-learning from kernel ridge-regression.
\newblock In \emph{International Conference on Learning Representations}, 2020.

\bibitem[Nguyen et~al.(2021)Nguyen, Novak, Xiao, and Lee]{nguyen2021dataset}
Nguyen, T., Novak, R., Xiao, L., and Lee, J.
\newblock Dataset distillation with infinitely wide convolutional networks.
\newblock \emph{Advances in Neural Information Processing Systems}, 34:\penalty0 5186--5198, 2021.

\bibitem[Sachdeva \& McAuley(2023)Sachdeva and McAuley]{sachdeva2023data}
Sachdeva, N. and McAuley, J.
\newblock Data distillation: A survey.
\newblock \emph{arXiv preprint arXiv:2301.04272}, 2023.

\bibitem[Schirrmeister et~al.(2022)Schirrmeister, Liu, Hooker, and Ball]{schirrmeister2022less}
Schirrmeister, R.~T., Liu, R., Hooker, S., and Ball, T.
\newblock When less is more: Simplifying inputs aids neural network understanding.
\newblock \emph{arXiv preprint arXiv:2201.05610}, 2022.

\bibitem[Shin et~al.(2024)Shin, Shin, and Moon]{shin2024frequency}
Shin, D., Shin, S., and Moon, I.-C.
\newblock Frequency domain-based dataset distillation.
\newblock \emph{Advances in Neural Information Processing Systems}, 36, 2024.

\bibitem[Simonyan \& Zisserman(2015)Simonyan and Zisserman]{simonyan2015very}
Simonyan, K. and Zisserman, A.
\newblock Very deep convolutional networks for large-scale image recognition.
\newblock In \emph{3rd International Conference on Learning Representations (ICLR 2015)}, 2015.

\bibitem[Vicol et~al.(2022)Vicol, Lorraine, Pedregosa, Duvenaud, and Grosse]{vicol2022implicit}
Vicol, P., Lorraine, J.~P., Pedregosa, F., Duvenaud, D., and Grosse, R.~B.
\newblock On implicit bias in overparameterized bilevel optimization.
\newblock In \emph{International Conference on Machine Learning}, pp.\  22234--22259. PMLR, 2022.

\bibitem[Wang et~al.(2023)Wang, Adebayo, Tan, Garcia-Olano, and Kokhlikyan]{wang2023error}
Wang, F., Adebayo, J., Tan, S., Garcia-Olano, D., and Kokhlikyan, N.
\newblock Error discovery by clustering influence embeddings.
\newblock In \emph{NeurIPS}, 2023.

\bibitem[Wang et~al.(2022)Wang, Zhao, Peng, Zhu, Yang, Wang, Huang, Bilen, Wang, and You]{wang2022cafe}
Wang, K., Zhao, B., Peng, X., Zhu, Z., Yang, S., Wang, S., Huang, G., Bilen, H., Wang, X., and You, Y.
\newblock Cafe: Learning to condense dataset by aligning features.
\newblock In \emph{Proceedings of the IEEE/CVF Conference on Computer Vision and Pattern Recognition}, pp.\  12196--12205, 2022.

\bibitem[Wang et~al.(2018)Wang, Zhu, Torralba, and Efros]{wang2018dataset}
Wang, T., Zhu, J.-Y., Torralba, A., and Efros, A.~A.
\newblock Dataset distillation.
\newblock \emph{arXiv preprint arXiv:1811.10959}, 2018.

\bibitem[Wu et~al.(2023)Wu, Deng, and Russakovsky]{wu2023multimodal}
Wu, X., Deng, Z., and Russakovsky, O.
\newblock Multimodal dataset distillation for image-text retrieval.
\newblock \emph{arXiv preprint arXiv:2308.07545}, 2023.

\bibitem[Yao et~al.(2020)Yao, Gholami, Keutzer, and Mahoney]{yao2020pyHessian}
Yao, Z., Gholami, A., Keutzer, K., and Mahoney, M.~W.
\newblock Pyhessian: Neural networks through the lens of the hessian.
\newblock In \emph{2020 IEEE international conference on big data (Big data)}, pp.\  581--590. IEEE, 2020.

\bibitem[Zhao \& Bilen(2021)Zhao and Bilen]{zhao2021dataset}
Zhao, B. and Bilen, H.
\newblock Dataset condensation with differentiable siamese augmentation.
\newblock In \emph{International Conference on Machine Learning}, pp.\  12674--12685. PMLR, 2021.

\bibitem[Zhao \& Bilen(2023)Zhao and Bilen]{zhao2023dataset}
Zhao, B. and Bilen, H.
\newblock Dataset condensation with distribution matching.
\newblock In \emph{Proceedings of the IEEE/CVF Winter Conference on Applications of Computer Vision}, pp.\  6514--6523, 2023.

\bibitem[Zhao et~al.(2021)Zhao, Mopuri, and Bilen]{zhao2020dataset}
Zhao, B., Mopuri, K.~R., and Bilen, H.
\newblock Dataset condensation with gradient matching.
\newblock In \emph{Ninth International Conference on Learning Representations 2021}, 2021.

\bibitem[Zhao et~al.(2023)Zhao, Li, Qin, and Yu]{zhao2023improved}
Zhao, G., Li, G., Qin, Y., and Yu, Y.
\newblock Improved distribution matching for dataset condensation.
\newblock In \emph{Proceedings of the IEEE/CVF Conference on Computer Vision and Pattern Recognition}, pp.\  7856--7865, 2023.

\bibitem[Zhong \& Liu(2023)Zhong and Liu]{zhong2023towards}
Zhong, X. and Liu, C.
\newblock Towards mitigating architecture overfitting in dataset distillation.
\newblock \emph{arXiv preprint arXiv:2309.04195}, 2023.

\bibitem[Zhou et~al.(2022)Zhou, Nezhadarya, and Ba]{zhou2022dataset}
Zhou, Y., Nezhadarya, E., and Ba, J.
\newblock Dataset distillation using neural feature regression.
\newblock \emph{Advances in Neural Information Processing Systems}, 35:\penalty0 9813--9827, 2022.

\end{thebibliography}
\bibliographystyle{icml2024}

 \newpage
 \appendix
 \onecolumn
 \section{Loss Landscape Analysis}
 \label{sec:add_loss_analysis}
 In this section, we aim to provide more intuition on the behavior and information of distilled data through visualization of the loss landscape induced by distilled data. We also provide additional curvature analysis to better understand the nature of distilled data.
 \begin{figure*}[t]
    \centering
    \includegraphics[width=0.9\linewidth]{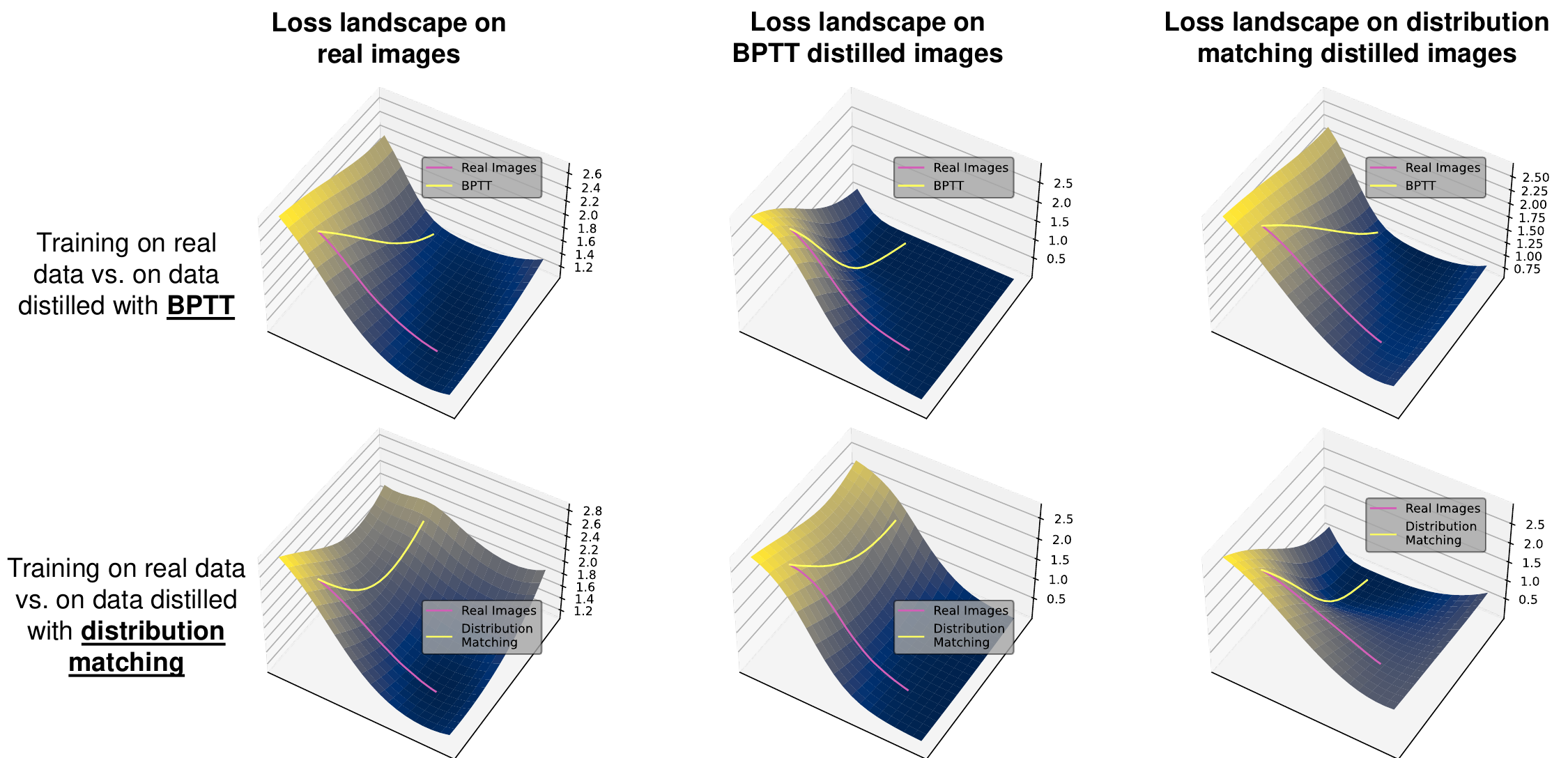}
    \caption{\textbf{Visualization of loss landscapes.} The loss landscapes evaluated with different data sources (columns) on trajectories of models trained on real data vs. models trained on data distilled with BPTT or distribution matching (rows). Two intriguing observations emerges. First, the BPTT landscape is smooth, leading to easy optimization, and flat in many places, shown in BPTT landscape when compared to test landscape for both BPTT and distribution matching trajectories. Second, landscape of distribution matching contains area of poor alignment with landscape of the real data shown in training trajectory on distribution matching distilled data.}
    \label{fig:loss}
    \vspace{1ex}
\end{figure*}
\smallsec{Visualization} The loss landscape of distilled data and real data can be visualized by generating a surface with two directional vectors $\delta$ and $\eta$ \cite{li2018visualizing}. Instead of using random vectors, which is good for analyzing convexity of loss landscape, we utilize the endpoints from training a model on different data sources to compare and contrast the training trajectories with respect to the loss landscape \cite{goodfellow2014qualitatively}. In more detail, we use a randomly initialized ConvNet with parameters $\theta_0$ to train the model with real data for one epoch, arriving at parameters $\theta_r$, and to train the same randomly initialized model with distilled data, arriving at parameters $\theta_d$. Afterwards, we set the two directional vectors $\delta = \theta_r - \theta_0$ and $\eta = \theta_d - \theta_0$. For better visualization, we use orthogonal component of $\eta$ to $\delta$ and a slight offset is applied. Finally, the two directional vectors specifies a hyperplane that allows us to sample the loss landscape. This procedure gives us the exact loss on the start and endpoint of the training trajectory with the rest defined as linear interpolation between the start and endpoints, giving us a set of loss values: $$L = \{\theta_0 + a*\delta + b*\eta \, | \, a,b \in [0, 1] \}.$$ 

We visualize the respective landscape on the training trajectory of models trained on real data and models trained on distilled data in Figure \ref{fig:loss}. We reveal two distinct patterns between BPTT and distribution matching. Training on real data, the loss landscape of BPTT on the model tend to quickly converges to a very flat region. As a consequence, the model quickly converges and no additional learning occurs, which is akin to early stopping. Additionally, this also explains the high predictive accuracy across models shown in Figure \ref{fig:model} as the high performing models land in the flat region. In contrast, landscape of distribution matching contains area of poor alignment with test landscape as shown in Figure \ref{fig:loss}, leading to poor performance in model training as well as generalization shown in \ref{fig:model}. 
 
\begin{figure*}[h]
     \centering
     \includegraphics[width=0.99\linewidth]{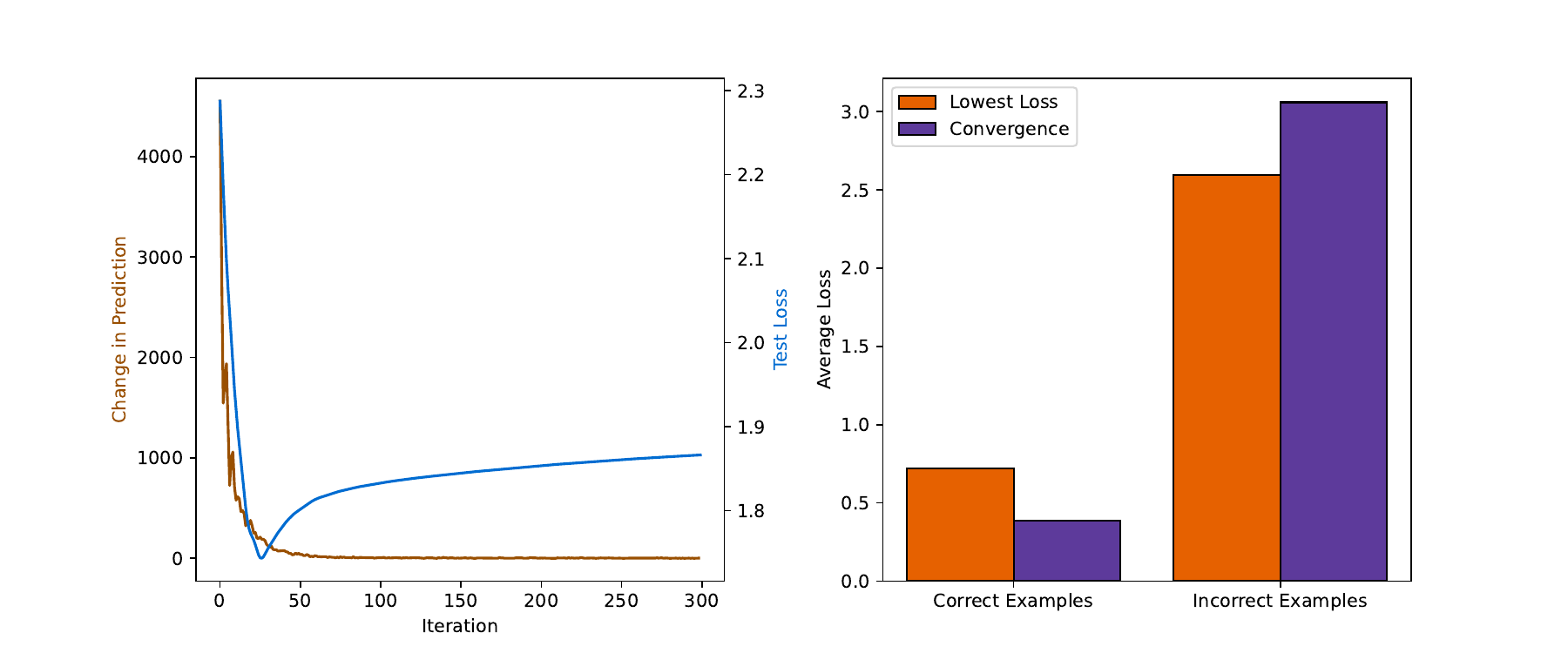}
     \caption{\textbf{Explanation for increase in test loss on distribution matching.} \textit{left.} The change in model prediction on test examples and the average loss on the test set. We observe that while there is little change in prediction after iteration 50, the test loss steadily increases. \textit{right.} We breakdown the loss on the intermediate model with the lowest overall test loss and the final converged model by test examples which the final converged model get correctly. We reveal that increase in loss is nothing more than overconfidence on incorrect examples.}
     \label{fig:high_loss}
     
 \end{figure*}

 \smallsec{Loss increase in distribution matching} Figure \ref{fig:loss} presents an interesting observation where optimizing on distribution matching distilled data ends up increase in the loss on the test dataset. The observation raises question to the nature of this phenomenon: is the model getting worse or is the model getting more confident? A quick investigation suggests the latter. In Figure \ref{fig:high_loss} \textit{left}, we perform a simple sanity check to check if the model predictions are changing. We reveal in the figure that the actual prediction is not changing much after 50 iteration. The loss, however, does increase even when the prediction does not change. In Figure \ref{fig:high_loss} \textit{right}, we divide the loss calculation by the examples on whether the final converged models get it correctly. We perform this calculation on two models: the intermediate model with the lowest test loss and the final converged model. We reveal in Figure \ref{fig:high_loss} \textit{right} that the model with the lowest overall test loss has higher loss on the correct examples but lower loss on the incorrect examples. Since distribution matching result in classification accuracy less than 50\%, the overconfidence in incorrect examples leads to a higher overall loss, explaining the increase in test loss. 

\begin{figure*}[t]
    \centering
    \includegraphics[width=0.99\linewidth]{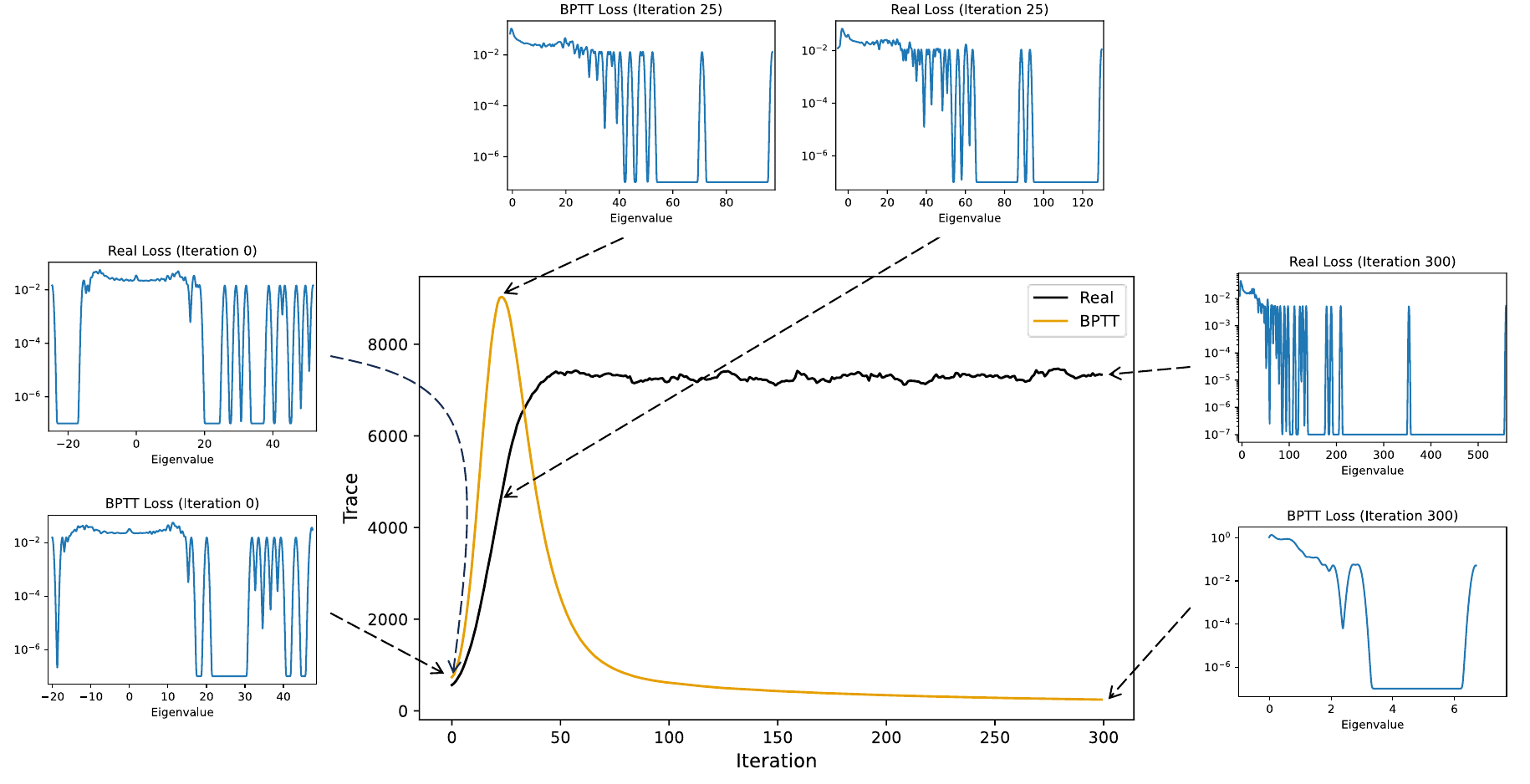}
    \caption{\textbf{Curative of loss landscape when trained on BPTT.} The plot shows the (smoothed) trace of the Hessian matrix on BPTT distilled data (what the model is trained on) and real data (actual loss landscape). Additionally, a more detailed breakdown of eigenvalues is shown with log density plots of eigenvalues at specific iteration of interest is shown. We observe that the model arrives at a flat region of the BPTT loss landscape relatively quickly in less than 100 iterations, but the actual loss landscape is converged at a very sharp region with a couple of very high eigenvalues shown in the log density plot.}
    \label{fig:trace_BPTT}
    \vspace{1ex}
\centering
\includegraphics[width=0.99\linewidth]{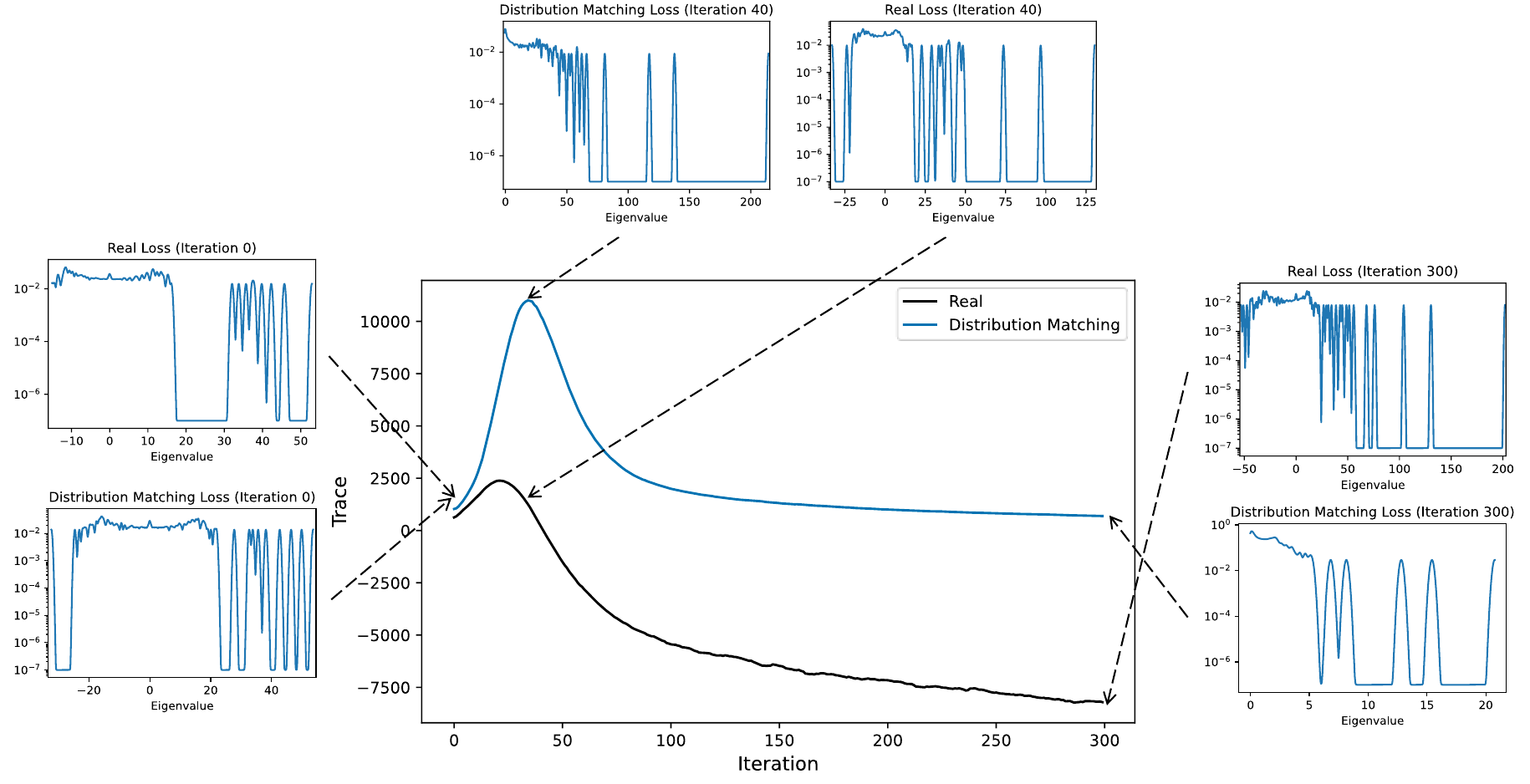}
\caption{\textbf{Curvature of loss landscape when trained on distribution matching.} The plot shows the (smoothed) trace of the Hessian matrix of distribution matching distilled data(what the model is trained on) and real data (actual loss landscape). Similar to Figure \ref{fig:trace_BPTT}, the log density of eigenvalues are specific iterations is also shown. Similar to BPTT, optimizing on the loss landscape of distribution matching also arrives at a very sharp region. However, we observe that the real loss landscape at iteration 40 and 300 is composed of very large number of negative eigenvalues, higher than what is observed initially at iteration 0.}
\label{fig:trace_DM}
\end{figure*}

 \smallsec{Dynamics of training on distilled data} We extend our curvature analysis from section \ref{sec:info_distilled} to the optimization trajectory of models trained on data distilled with BPTT and distribution matching. Figure \ref{fig:trace_BPTT} and \ref{fig:trace_DM} reveal the trace of the Hessian matrix of the landscape of the distilled data, what it is optimized on, and the landscape of the real data, the actual landscape of interest, at each training iteration. Additionally, we provide more fine-grained analysis on the log density plot of the eigenvalues at three points of interest: random initialization at start of training, the iteration with the highest trace on the distilled data landscape (iteration 25 for BPTT or iteration 40 for distribution matching), and end of training at iteration 300. 
 
 In both cases, optimizing over BPTT and distribution matching, the log density plot of the eigenvalues reveals that while both models converge towards a flat region in the distilled data landscape, the actual landscape is a very sharp region that is composed of many large eigenvalues. In particular in Figure \ref{fig:trace_DM}, we note that the optimization trajectory from learning on distribution matching actually results in a region with large number of large negative eigenvalues, revealing that optimizing on distribution matching distilled data never leave the saddle-point but rather only arrive at a sharper saddle-point. The increase of negative eigenvalues coupled with our analysis on the loss landscape in Figure \ref{fig:high_loss} where optimizing on distribution matching distilled data actually results in \emph{higher} loss suggests that there is a significant misalignment in loss landscape between real data and distribution matching distilled data and further optimizing could cause the model to move up the saddle point. 

\clearpage

\begin{figure}[h]
    \centering
    \begin{minipage}[t]{0.49\linewidth}
        \centering
        \includegraphics[width=\linewidth]{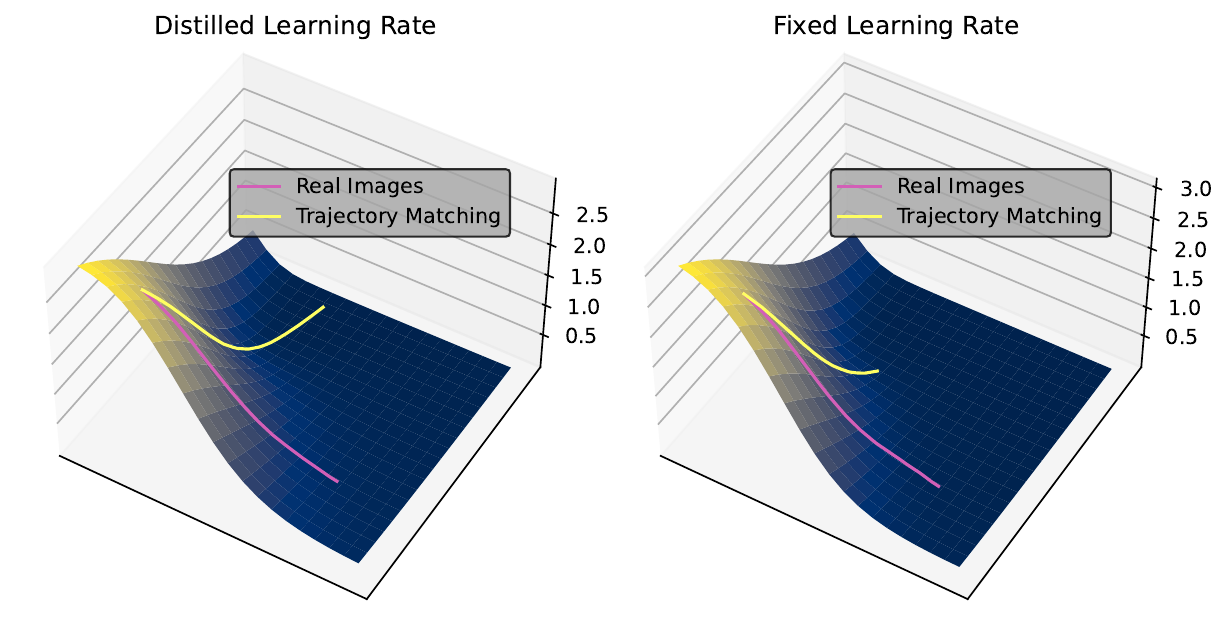}
    \end{minipage}
    \hfill
    \begin{minipage}[t]{0.40\linewidth}
        \centering
        \includegraphics[width=\linewidth]{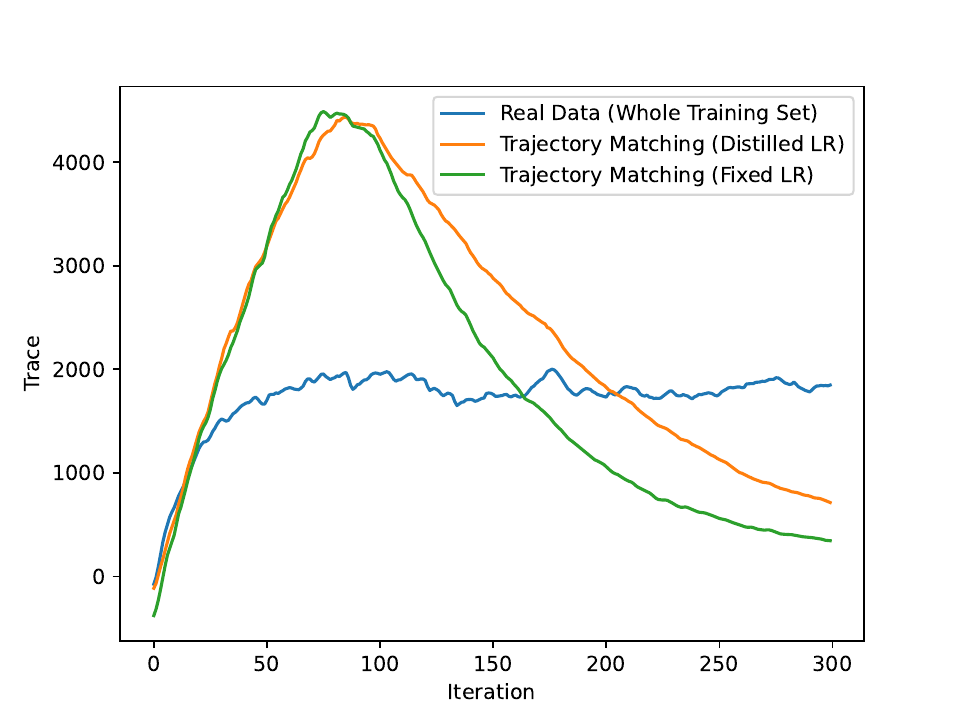}
    \end{minipage}
    \caption{\textbf{Effect of distilling learning rate in trajectory matching on loss landscape.} \textit{left.} Visualization of loss landscape of the distilled data that is analogous to Figure \ref{fig:loss} on a model trained on real data vs. trajectory matching data distilled with fixed or learned learning rates. The visualization reveals that whether or not to distill the learning rate does not change our findings in Figure \ref{fig:loss}: loss landscape of distilled data is smooth and easily optimizable. \textit{right.} Trace of the Hessian of a model trained on real data on the loss landscape of real data vs. trajectory matching data distilled with fixed or learned learning rates (analogous to Figure \ref{fig:eigen}). The trace more rigorously support our flatness argument: we observe that regardless whether learning rate is distilled, trajectory matching produces an easily optimizable path. The non-flat region of the loss landscape when the learning rate is distilled extends slightly further, corresponding to the higher information content i.e. better classification accuracy when trained on.}
    \label{fig:learningrate}
\end{figure}

\smallsec{Effects of distilling learning rates} We first analyze the effect of having a learnable learning rate during distillation process of trajectory matching shown in Figure \ref{fig:learningrate}. The curvature of the distilled data is similar regardless to whether learning rate is learned. There is a slight increase in the non-flat regions of the loss landscape when the learning rate is learned, which is consistent to the higher performing classification model when trained on the distilled data with the learned learning rate. 

\begin{figure}[h]
    \centering
    \begin{minipage}[c]{0.42\linewidth}
        \centering
        \includegraphics[width=\linewidth]{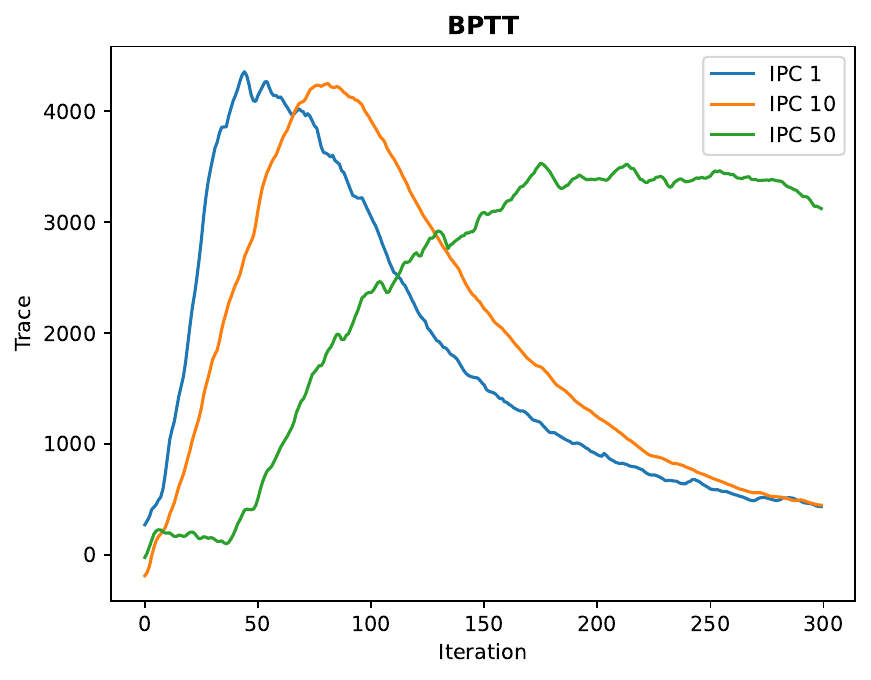}
    \end{minipage}
    \hfill
    \begin{minipage}[c]{0.42\linewidth} 
        \centering
        \includegraphics[width=\linewidth]{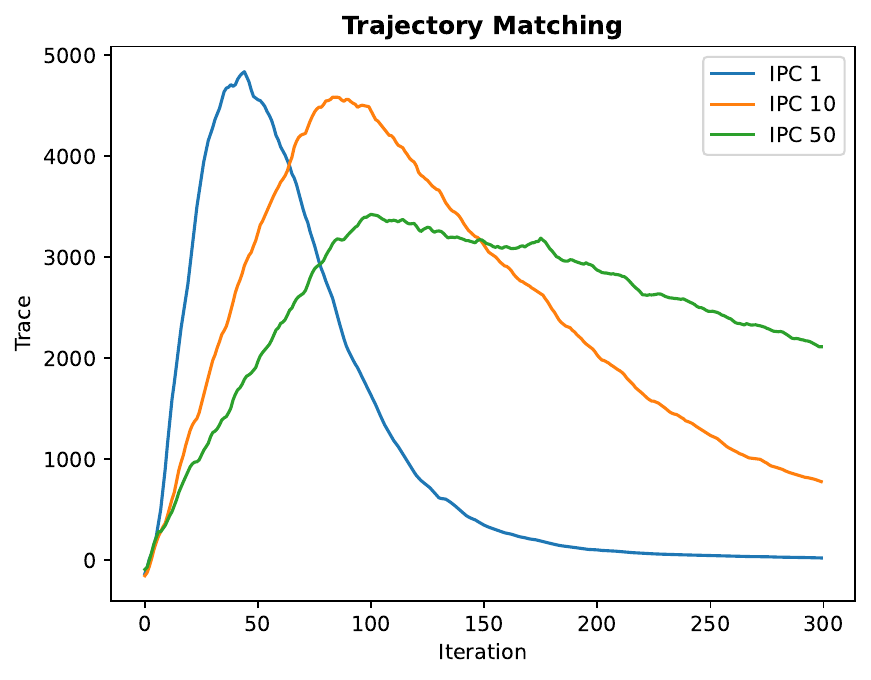}
    \end{minipage}
    \caption{\textbf{Curvature findings in Figure \ref{fig:eigen} generalizes to different number of images per class (IPC).} \textit{left.} Trace of the Hessian matrix (smoothed) evaluated on data distilled with BPTT at every iteration of a model trained on real data for 300 iterations. We observe that our conclusion on distilled data is consistent and captures early dynamics of training that gradually shift towards later parts of the training as we increase the IPC, which also explains the performance increase with higher IPC. \textit{right.} The same experimental setup as the left figure but using data distilled with Trajectory Matching. We observe the same consistent pattern that supports our conclusion that distilled data captures early training dynamics.}
    \label{fig:ipc_scale}
\end{figure}

\smallsec{Effect of storage budget} We also analyze the effect of the budget, different number of images per class (IPC), on the curvature of the distilled loss landscape with BPTT and trajectory matching. We observe in Figure \ref{fig:ipc_scale} that that with low budget (1 IPC), the trace of the Hessian quickly peaks at around iteration 50 while high budget (50 IPC) peaks after iteration 100 with a more gradual decrease. Such findings reveal that our analysis in Figure \ref{fig:eigen} generalizes to the increase in budget: more distilled images capture more of the training trajectory, and hence, higher performance.

\clearpage

 \section{Additional Mixing and Prediction Analysis}
 \label{sec:add_pred_analysis}
 In this section, we detail additional analyses on the mixing experiment done in Figure \ref{fig:data} \textit{right} as well prediction analysis done in Figure \ref{fig:agreement} to show our findings generalizes to different datasets, larger scales, and state-of-the-art (SOTA) methods.

\begin{figure}[h]
    \vspace{1.0em}
    \centering
    \begin{minipage}[c]{0.49\linewidth}
        \centering
        \includegraphics[width=\linewidth]{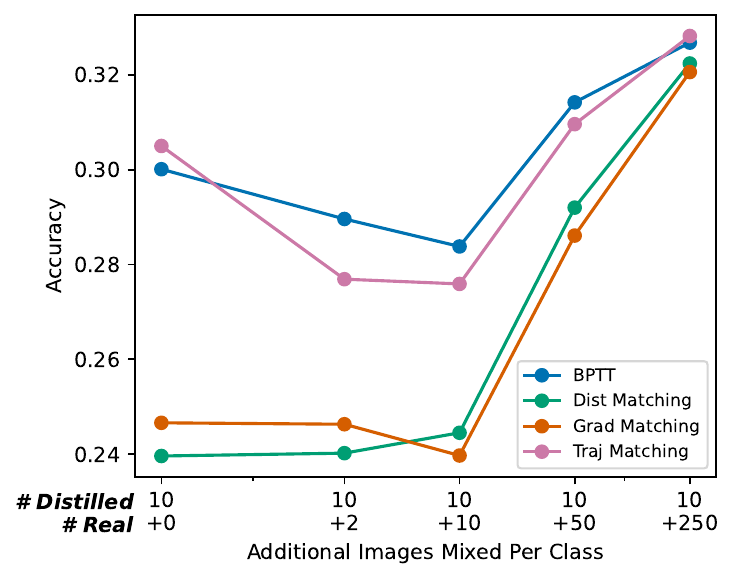}
    \end{minipage}
    \hfill
    \begin{minipage}[c]{0.44\linewidth}
        \vspace{-1.9em}
        \centering
        \includegraphics[width=\linewidth]{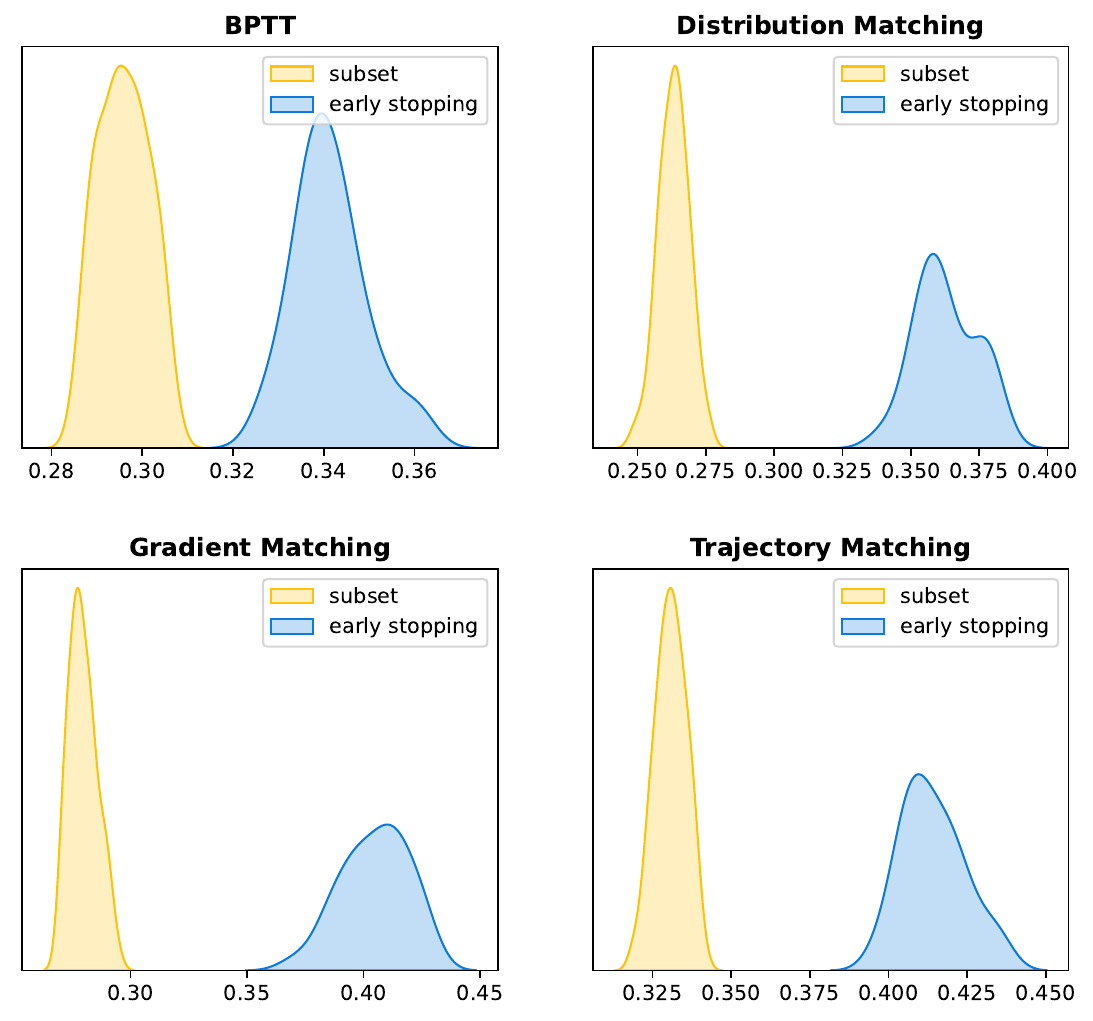}
    \end{minipage}
    \caption{\textbf{Main conclusions generalize to CIFAR-100.} \textit{left.} Accuracy of models trained on distilled data mixed with real data analogous to Figure \ref{fig:data} but on CIFAR-100 dataset. The result is consistent with our findings on the four baselines: mixing real data with distilled data does not necessarily lead to performance increase and can sometime even cause performance drop. \textit{right.} Kernel density estimation (KDE) plots on the number of test data points where predictions of distilled-trained models agree with early-stopped/subset-trained models, similar to Figure \ref{fig:agreement} but using the CIFAR-100 dataset. The finding is consistent with our main conclusion where distilled-trained models have higher agreement on a test data's prediction with early-stopped models than subset-trained models.}
    \label{fig:cifar100}
\end{figure}

\smallsec{CIFAR-100 experiments} We extend our analysis on mixing with real data and the model predictions analysis to the CIFAR-100 dataset \cite{krizhevsky2009learning}. We observe similar trends in Figure \ref{fig:cifar100} \textit{left} where addition of real data causes performance increases. Additionally, density plots from Figure \ref{fig:cifar100} \textit{right} finds the same consistent finding: predictions of models trained on distilled data tend to agree more with models trained on whole train dataset that is stopped early rather than models trained on a subset of data. In consequence, this analysis reveal that the patterns observed in Figure \ref{fig:data} and Figure \ref{fig:agreement} is not dataset specific and extends beyond CIFAR-10. 

\begin{figure}[t]
    \centering
    \resizebox{0.85\linewidth}{!}{ 
    \begin{minipage}[c]{0.39\linewidth}
        \centering
        \includegraphics[width=\linewidth]{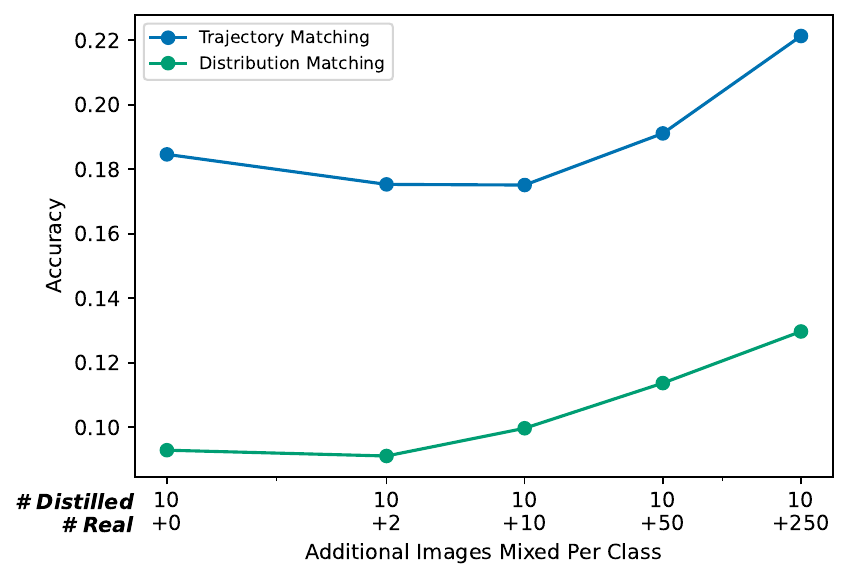}
    \end{minipage}
    \hfill
    \begin{minipage}[c]{0.55\linewidth}
        \vspace{-1.9em} 
        \centering
        \includegraphics[width=\linewidth]{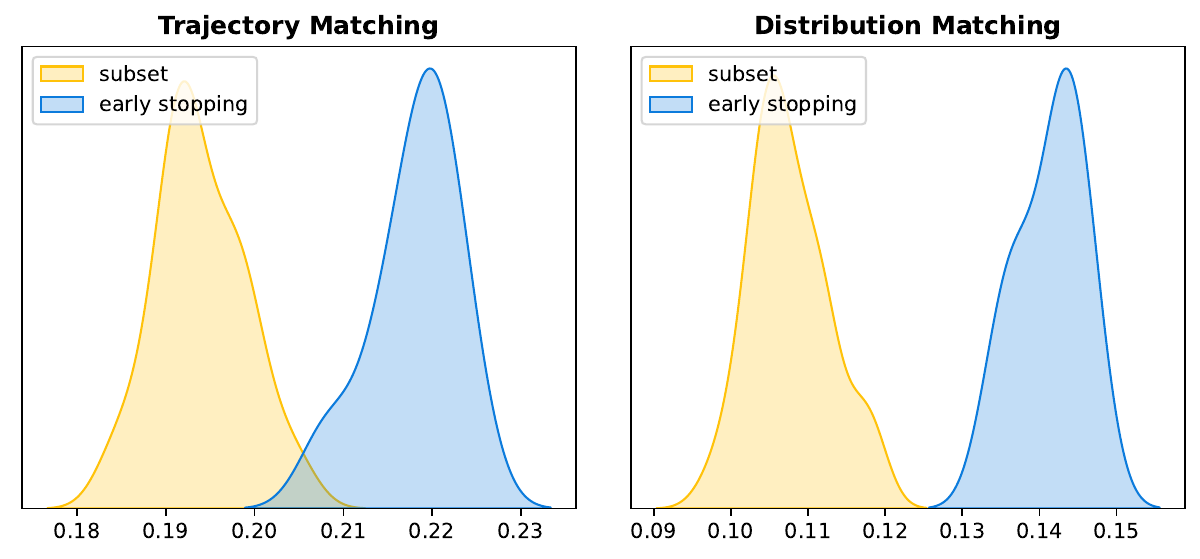}
    \end{minipage}
    }
    \caption{\textbf{Main conclusions generalize to the TinyImageNet dataset.} \textit{left.} Accuracy of models trained on distilled data mixed with real data analogous to Figure \ref{fig:data} but on the higher resolution TinyImageNet dataset. BPTT and Gradient Matching are excluded because the algorithm fails to converge to a reasonable solution with random chance final classification accuracy. The result is consistent with our findings on CIFAR-10: mixing real data with distilled data does not necessarily lead to performance increase and can sometime even cause performance drop. \textit{right.} KDE plots on the number of test data points where predictions of distilled-trained models agree with early-stopped/subset-trained models, similar to Figure \ref{fig:agreement} but on the higher resolution TinyImageNet dataset. The finding is consistent with our main conclusion where distilled-trained models have higher agreement on a test data's prediction with early-stopped models than subset-trained models.}
    \label{fig:tinyimagenet}
\end{figure}

\smallsec{TinyImageNet experiments} Similarly, we extend mixture experiments and our model prediction analysis to the TinyImageNet dataset \cite{le2015tiny}, which composes images with higher resolution at 64x64. Mixture accuracy and density plots shown in Figure \ref{fig:tinyimagenet} reveal a similar trends, demonstrating that the pattern observed in Figure \ref{fig:data} and Figure \ref{fig:agreement}  also scales with image resolution. 

\begin{figure}[t]
    \centering
    \resizebox{0.85\linewidth}{!}{ 
    \begin{minipage}[c]{0.49\linewidth}
        \centering
        \includegraphics[width=\linewidth]{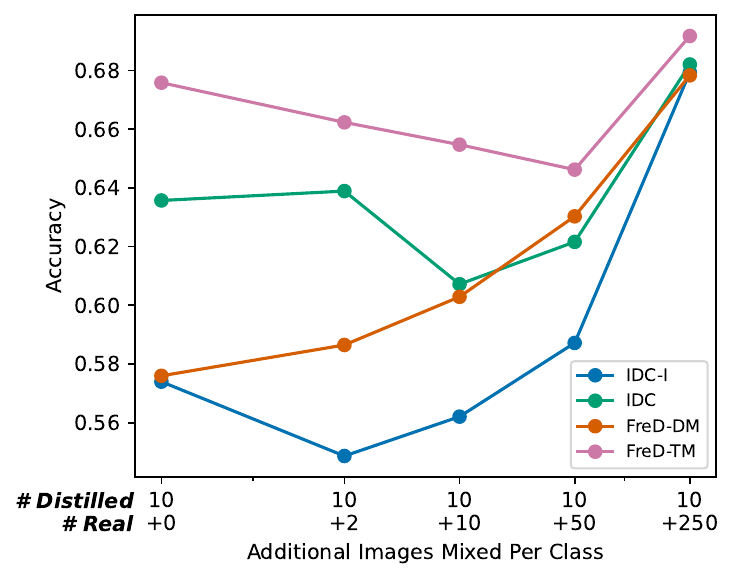}
    \end{minipage}
    \hfill
    \begin{minipage}[c]{0.44\linewidth}
        \vspace{-1.9em}
        \centering
        \includegraphics[width=\linewidth]{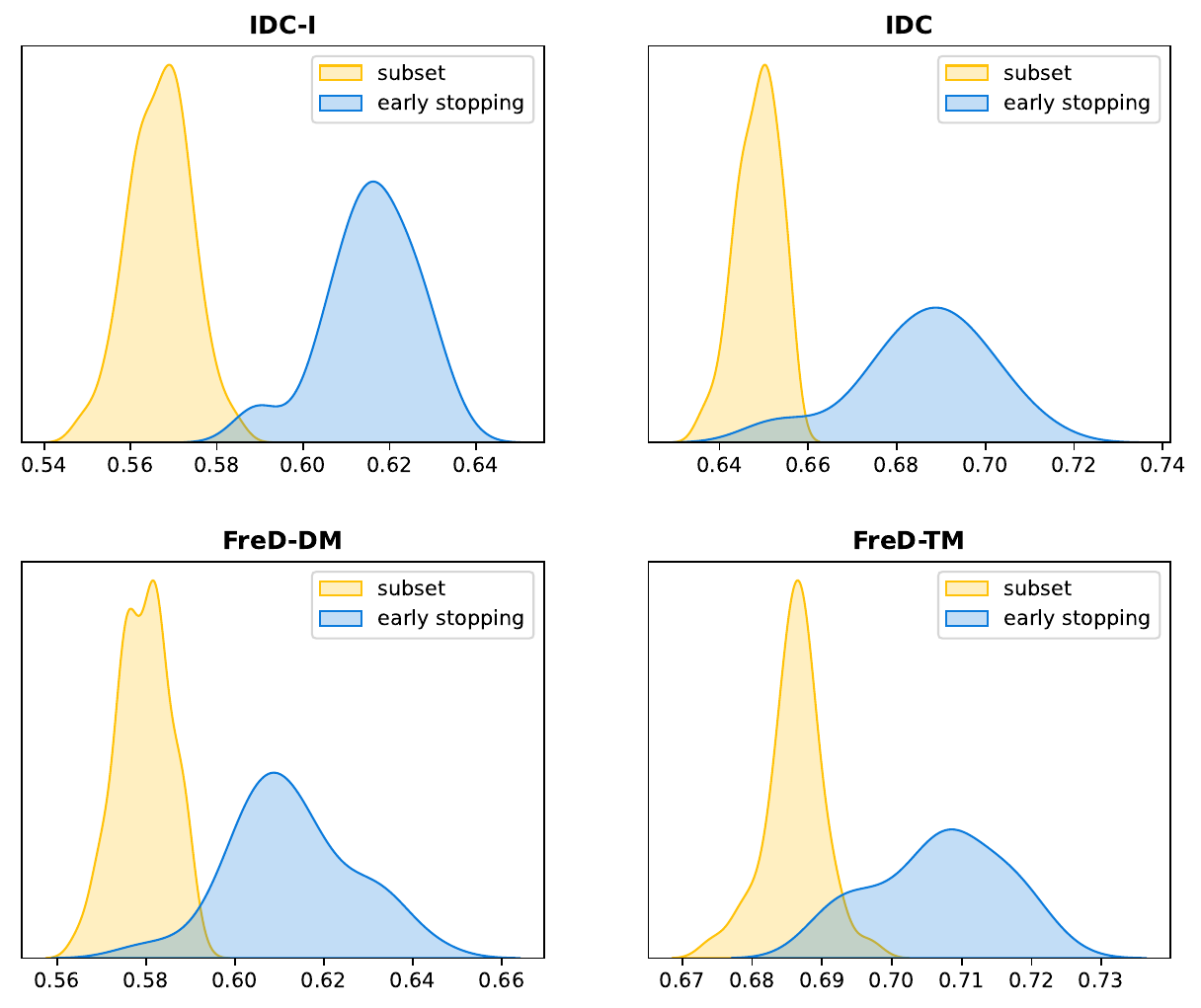}
    \end{minipage}
    }
    \caption{\textbf{Main conclusions generalize to the state-of-the-art (SOTA) dataset distillation methods.} \textit{left.} Accuracy of models trained on distilled data mixed with real data analogous to Figure \ref{fig:data} but using SOTA distillation methods. The result is consistent with our findings on the four baselines: mixing real data with distilled data does not necessarily lead to performance increase and can sometime even cause performance drop. \textit{right.} KDE plots on the number of test data points where predictions of distilled-trained models agree with early-stopped/subset-trained models, similar to Figure \ref{fig:agreement} but using SOTA distillation methods. The finding is consistent with our main conclusion where distilled-trained models have higher agreement on a test data's prediction with early-stopped models than subset-trained models.}
    \label{fig:sota}
\end{figure}

\smallsec{IDC and FreD experiments} We also extend mixture experiments and our model prediction analysis to state-of-the-art (SOTA) methods Information-intensive Dataset Condensation (IDC) \cite{kim2022dataset} and Frequency domain-based dataset Distillation (FreD) \cite{shin2024frequency}. Classification models trained on data distilled from IDC utilizes CutMix augmentation to achieve SOTA performance. Mixture accuracy and density plots shown in Figure \ref{fig:sota} reveal a similar trends, demonstrating that the pattern observed in Figure \ref{fig:data} and Figure \ref{fig:agreement} also extends beyond our selected baseline methods and to SOTA dataset distillation methods. 

\begin{figure}
    \centering
    \includegraphics[width=0.9\linewidth]{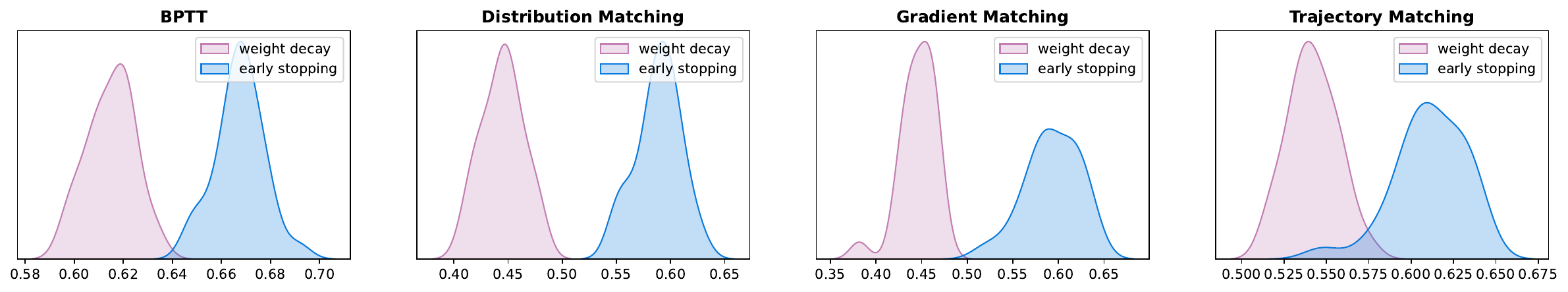}
    \caption{\textbf{Early-stopping is still more similar to distilled data than weight decay.} KDE plots on the number of test data points where predictions of distilled-trained models agree with early-stopped-models or models trained with decay. The early-stopped-models remains the most similar, further supporting our main conclusion. }
    \label{fig:weight_decay}
\end{figure}

\smallsec{Weight decay experiments} Lastly, we perform the identical prediction analysis done in Figure \ref{fig:agreement} but using models trained with weight decay in lieu of random subsets of training data. In a similar manner, the model is trained for 20 epochs to convergence. We train 300 models with weight decay where the strength is randomly selected 0.05 and 0.13 and selecting only models with similar test accuracy for comparisons. The density plot shown in Figure \ref{fig:weight_decay} reveal that early-stopped models is still more similar to distilled-trained-models than models trained with weight decay, furthering supporting our claim that distilled data is analogous to early-stopping.

\clearpage
\section{Variability of agreements}
\label{sec:variability}
\begin{figure}[t]
    \centering

    \includegraphics[width=0.70\linewidth]{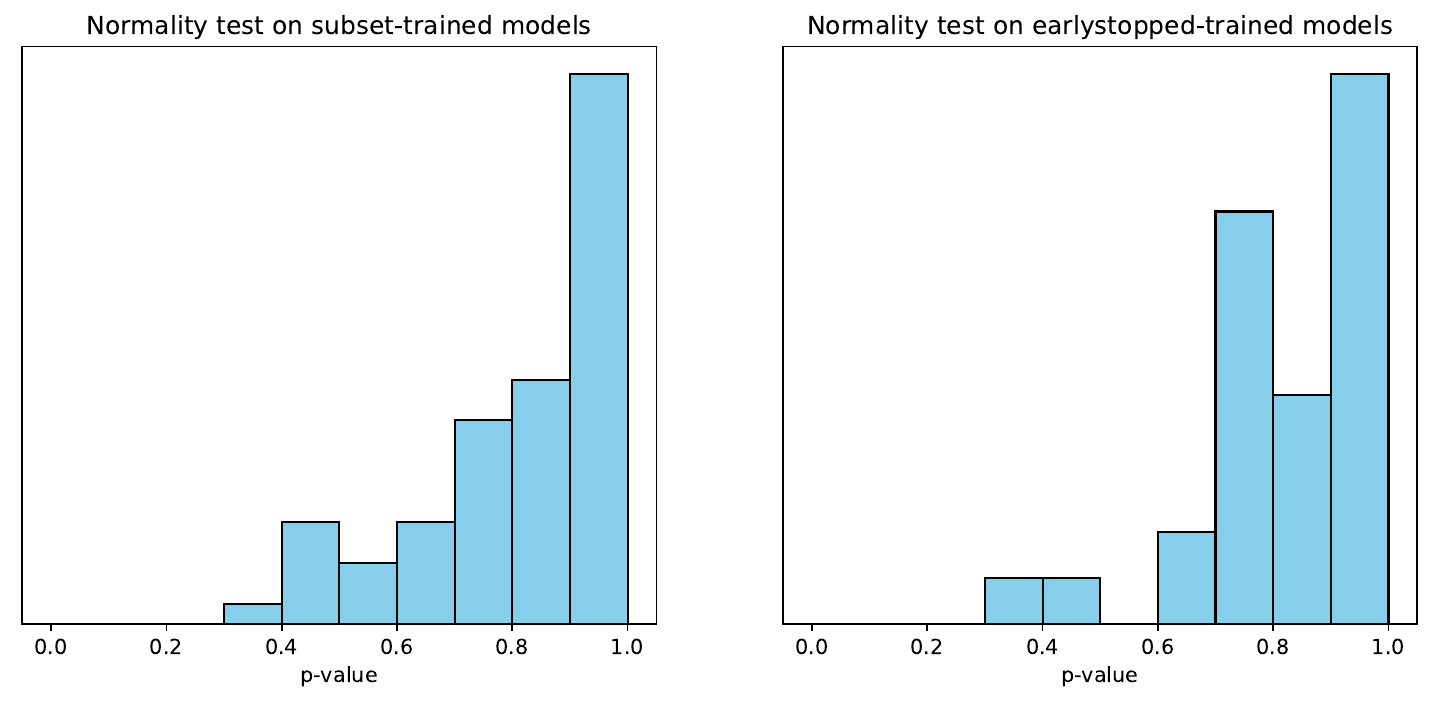}
    \caption{\textbf{Normality test on agreement predictions within subset-trained models and early-stopped models.} \textit{left.} Distribution of p-values of Kolmogorov Smirnov test against a normal distribution's cumulative density function on prediction agreement for every subset-trained models against every other subset-trained models. Non-significant p-values for every subset-trained models indicates failure to reject the null, and thus, suggests that distribution of agreement follows a normal distribution. \textit{right.} Same experimental setup as the left plot but with early-stopped models instead. The same conclusion is reached: distribution of agreement of early-stopped models with other early-stopped models follows a normal distribution.}
    \label{fig:pvalues}
\end{figure}

We quantitatively verify that the difference observed in Figure \ref{fig:agreement} cannot be attributed to the variability in agreement within the early-stopped or subset-trained models themselves. Similar to Figure \ref{fig:agreement}, our analysis was designed to be very fine-grained, focusing on each individual model rather than considering them in aggregate. To achieve this, we leveraged an interesting observation: the distribution of agreements between a particular subset-trained model and other subset-trained models follows a normal distribution. The same observation holds true for early-stopped models. We demonstrate using the Kolmogorov-Smirnov test \cite{lilliefors1967kolmogorov} against the cumulative density function (CDF) of a normal distribution shown in Figure \ref{fig:pvalues} where we fail to reject the null hypothesis, and hence, suggesting the distribution of agreements follows a normal distribution.

With this observation, we performed the following procedure:

\begin{enumerate}
    \item Start with a subset-trained model
    \item Obtain its agreement in predictions with every other subset-trained models
    \item Fit a normal distribution on the observed number of agreements
    \item Obtain the agreement between the distilled-trained model and the selected subset-trained model from step 1
    \item Calculate the probability of observing the agreement between the distilled-trained model and the selected subset-trained model using the CDF of the normal distribution
    \item Repeat step 1-5 with every other subset-trained model
\end{enumerate}

\begin{figure}[h]
    \includegraphics[width=\linewidth]{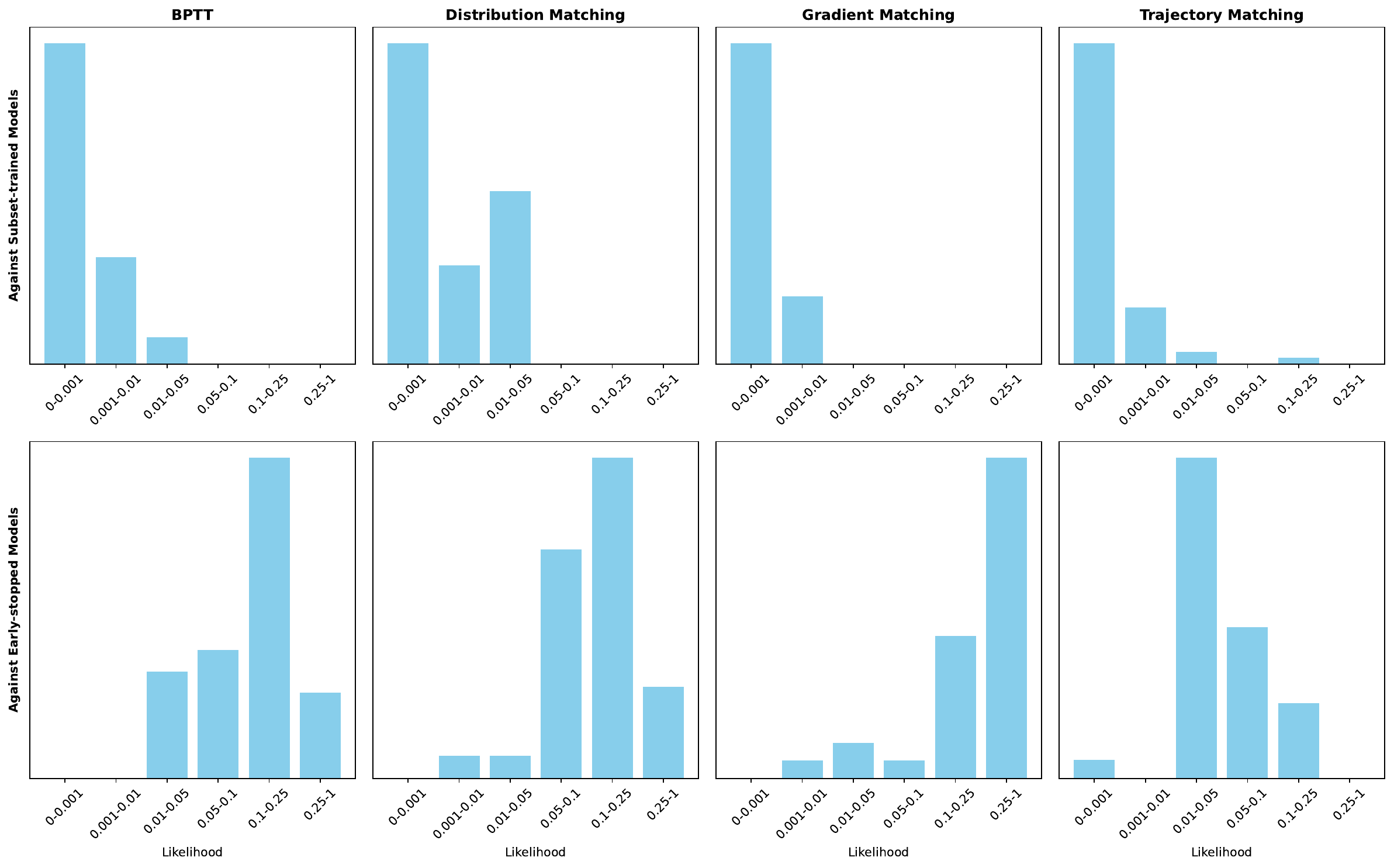}
    \caption{\textbf{Probability of observing the number of agreements in predictions with distilled-trained models vs. subset-trained models and early-stopped models.} The \textit{top} row reveals the probability of observing the number of agreements in the distilled-trained model vs. subset-trained-models given the variability in agreements within subset-trained models. The \textit{bottom} calculate the same probability but against early-stopped models and their variability. The results reveal lower probabilities of observing the number of agreements against distilled-trained-models for subset-trained models, supporting our findings in Figure \ref{fig:agreement}: distilled-trained models are more similar to early-stopped models.}
    \label{fig:likelihood}
\end{figure}

We perform the same procedure with early-stopped models in place of the subset-trained models as well. The resulting histogram between subset-trained models and early-stopped models in Figure \ref{fig:likelihood} shows a notable pattern: probabilities of observing the agreement when compared to subset-trained models is very low while probabilities of observing the agreement are significantly higher for early-stopped models. The finding supports the claim in the paper that distilled-trained models are more similar to early-stopped models than subset-trained models: i.e. the observed agreement from distilled-trained models can be explained away by variability of early-stopped models but not by variability of subset-trained models.

\clearpage

\section{Does distilled data points contain information outside of the labeled class?} 
\begin{figure}[t]
    \includegraphics[width=0.99\linewidth]{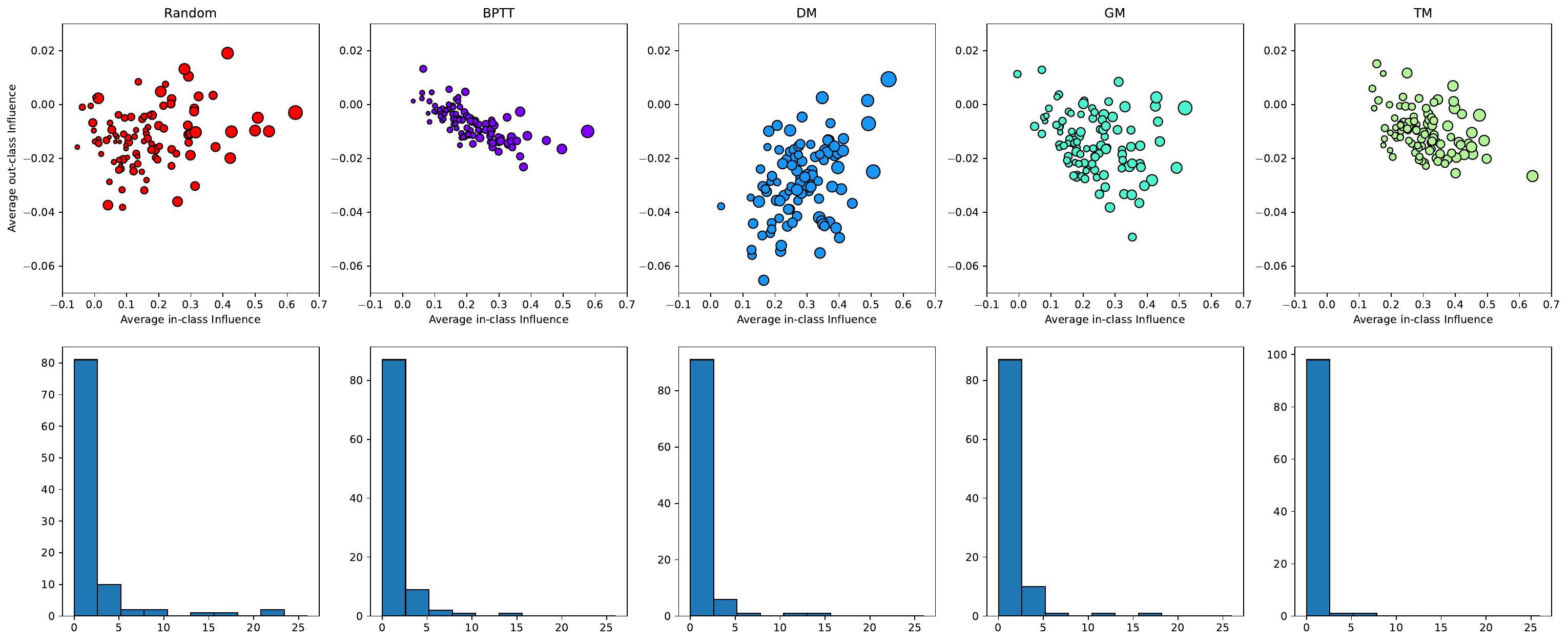}
    \caption{\textbf{Distilled data only contains information about the same class.} \textit{top.} For every image in a dataset, either the 100 distilled images or 100 random real images, we plot its average influence on all the test images with the same class (in-class) and all the test images with a different class (out-class). The variance in the influence across the whole test set is indicated by the size of the dot. The plots fail to reveal any distilled data point that has higher influence on out-class test images compared with randomly selected real images. \textit{bottom.} Using the same experimental setup but plotting the number of the top 100 most influenced test images that has a different class label also fails to reveal any indication that distilled images contain more information from other classes compared to real images.}
    \label{fig:outclass}
\end{figure}

A natural question of whether particular distilled data points provide information beyond its labeled class emerges from our analysis in Section \ref{sec:influence}. Our analysis below did not provide any strong evidence that distilled data behave any different from real data in this regard. In more detail, we performed analysis on the average influence of distilled data on in-class (same class as distilled data) test data and out-class (different class as distilled data) test data. The resulting Figure \ref{fig:outclass} \textit{top} reveals that, while there are distilled data points with positive out-class average influence, the actual quantity of the influence isn’t too different from what we expect with random real data. Additionally, we looked into the top 10 images with highest influence shown in Figure \ref{fig:outclass} \textit{bottom} and found none of the test images are of different classes. Extending this to top 100 images, there are several test images with different classes but the amount isn’t greater than what we expect with real images. In fact, there are fewer, with trajectory matching having the fewest, which is surprising. All these findings suggest the absence of strong signals that would indicate that distilled data is storing information about other classes. 

\clearpage

\section{Connecting Curvature with Information Content}
\begin{figure}
    \centering
    \includegraphics[width=0.70\linewidth]{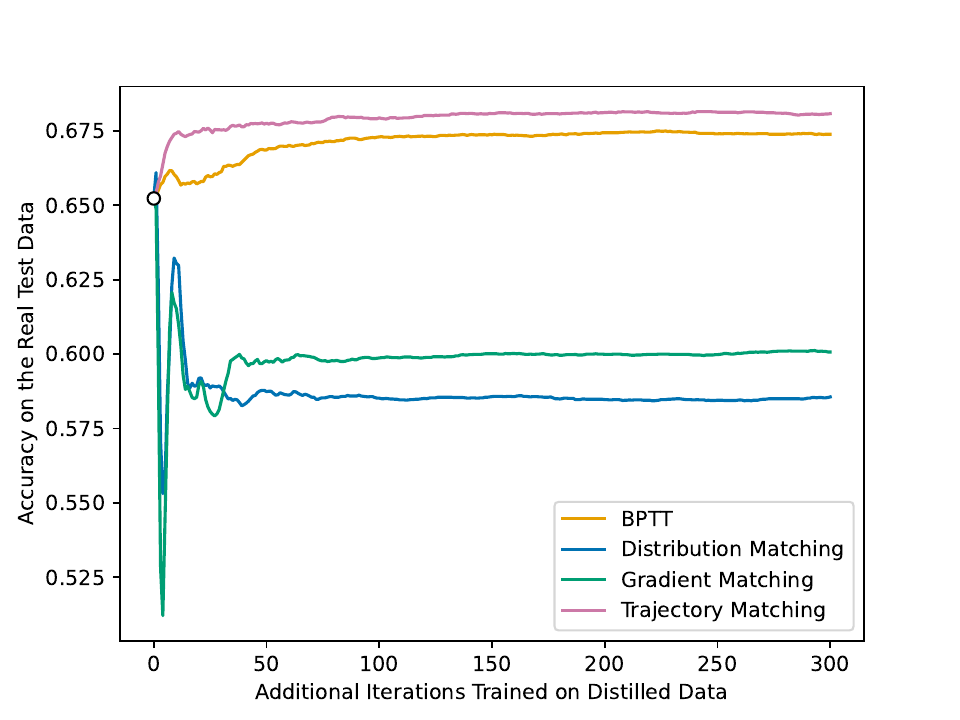}
    \caption{\textbf{Test accuracy changes from training on distilled data using a model pre-trained on real data for 300 iterations.} Utilizing a model that is trained on real data for 300 iterations (test accuracy is shown as white circle on the plot), we train the model on data distilled by BPTT, distribution matching, gradient matching, and trajectory matching for an additional 300 iterations. We reveal that additional training in BPTT/trajectory matching distilled data results in less than 2\% improvement on accuracy, confirming the flatness region from our curvature analysis. More importantly, we reveal that additional training on distribution matching and gradient matching distilled data results in decrease in test accuracy. This explains the high sharpness observed in our curvature analysis: while distribution matching and gradient matching distilled data do provide significant change to the model after the early iterations, the additional information stored can actually noise that is not pertinent to the classification task. }
    \label{fig:additional_training}
\end{figure}

We directly confirm our intuition on the relationship between the local curvature induced by distilled data and the information within distilled data. We explicit calculate the relevant information contained with distilled data by training a model that was pre-trained on real data for 300 iterations and calculating the task accuracy after the additional training. Figure \ref{fig:additional_training} confirms our intuition - additional training on BPTT and trajectory matching distilled data results in minimal changes in the accuracy of the model due to the flat region induced by the distilled data. Interestingly, the task accuracy also explains the high curvature induced by distribution matching and gradient matching distilled data observed in Figure \ref{fig:eigen}. Additional training on distribution matching and gradient matching distilled data does change the model, which align with the intuition of high curvature, but the change hurts the task performance rather than improve. Figure \ref{fig:additional_training} shows that additional training on distribution matching and gradient matching distilled data cause a 5-7 \% decrease in classification accuracy. 
\label{sec:additional_training}
\clearpage

\section{Initializing BPTT with Real Images}
\label{sec:bptt_init}
\begin{figure*}[t]
\centering
    \includegraphics[width=0.94\linewidth]{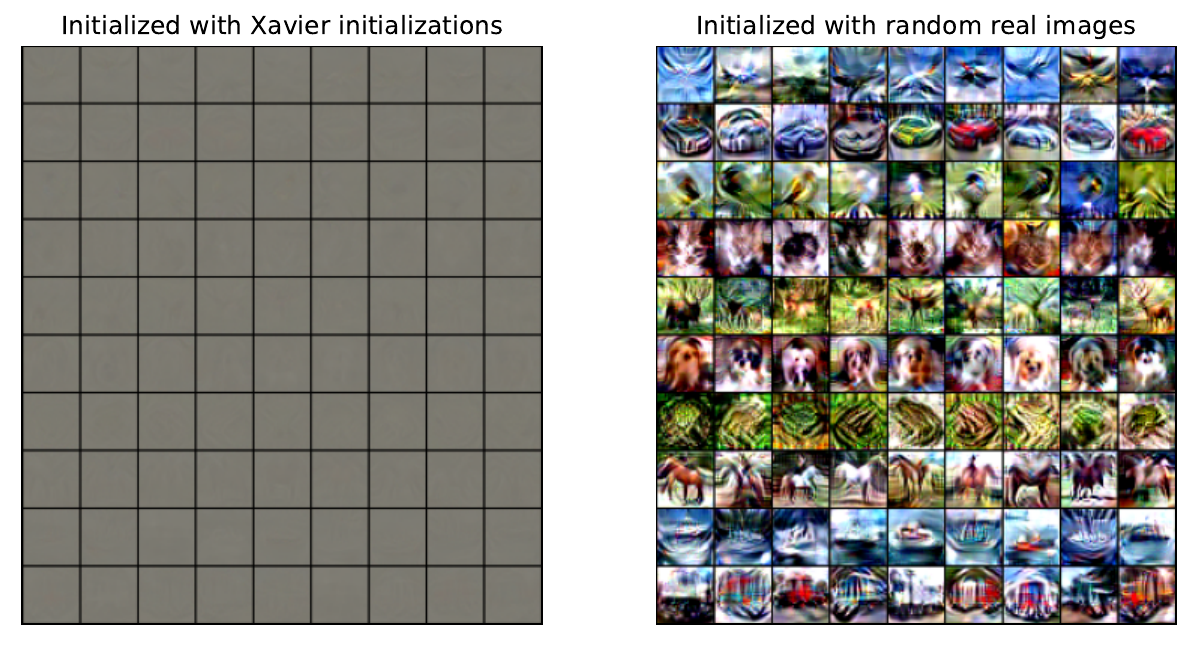} 
    \caption{\textbf{Visualization of BPTT distilled images with different initialization.} We observe that BPTT initialized with Xavier initialization converges to very gray looking images while randomly selecting real images as initialization produces distilled images with salient visual features similar to trajectory matching.}
    \label{fig:bptt_init}
\end{figure*}

\begin{figure*}[t]
    \includegraphics[width=0.49\linewidth]{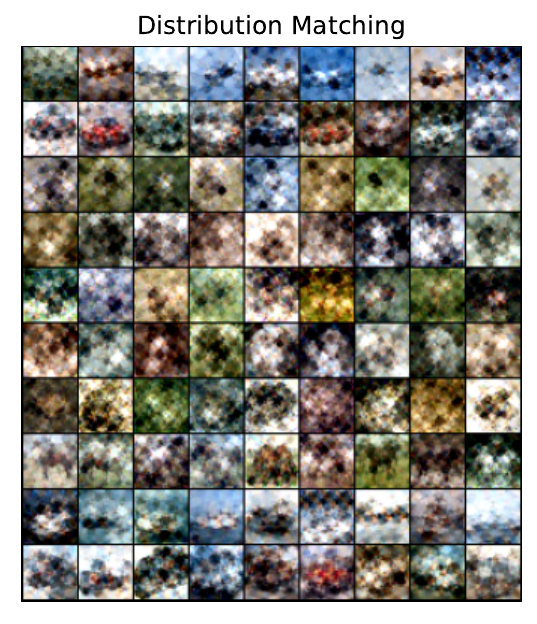}
    \includegraphics[width=0.49\linewidth]{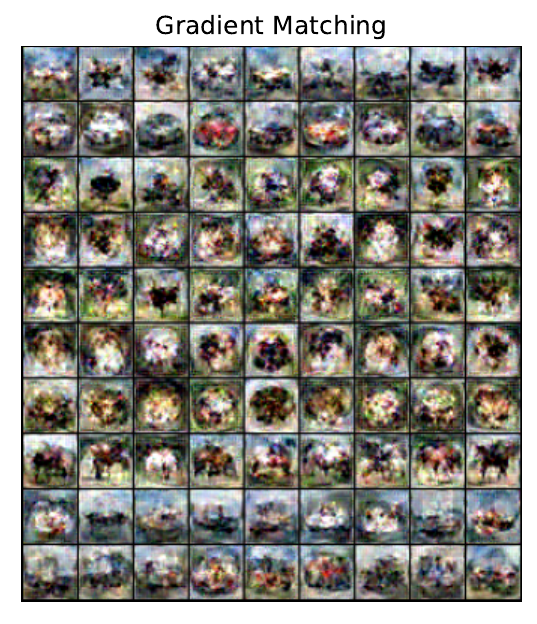}
    \caption{ \centering Visualization of data distilled by distribution matching and gradient matching.}
    \label{fig:DMGM_vis}
\end{figure*}

\begin{figure*}[t]
    \centering
    \includegraphics[width=0.49\linewidth]{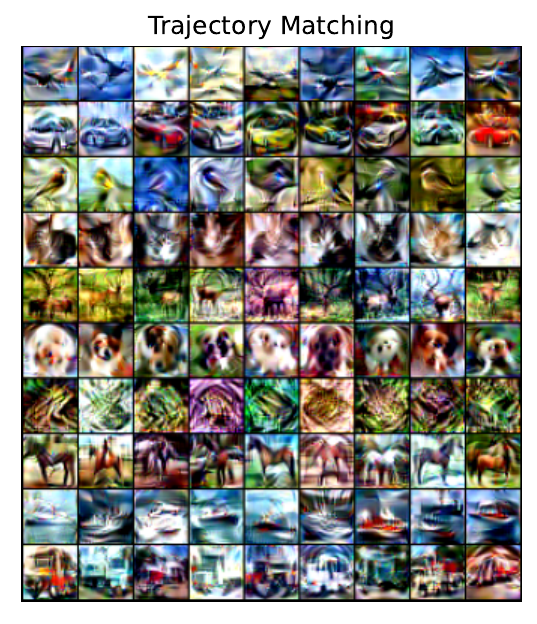}
    \caption{ \centering Visualization of data distilled by trajectory matching.}
    \label{fig:TM_vis}
\end{figure*}

We generate distilled data using BPTT for our analyses with minor but notable change compared to the original paper. The original proposed algorithm \cite{deng2022remember} utilizes Xavier initialization on distilled data since initialization did not impact final classification performance. While this is true, we found that Xavier initialization does not produce distilled images that is visually distinguishable and produce good behavior outside standard evaluation protocol (train a new model only on distilled data). In more detail, Figure \ref{fig:bptt_init} visually compares the 100 images distilled by BPTT using Xavier initialization with images distilled by BPTT using initialization from random real images; Xavier initialization produces very gray images, which is very unlike images distilled by distribution matching/gradient matching shown in Figure \ref{fig:DMGM_vis} and trajectory matching in Figure \ref{fig:TM_vis}. Meanwhile simply initialization with random real images produces distilled data that is more analogous to the other three studied distilled datasets. Additionally, we inspect the cross architecture generalization performance between the BPTT distilled data with Xavier initialization vs. random real images. To remove the effects of hyper-parameters, we perform 1000 iterations of random hyper-parameters search and compare the best accuracy. The result in Table \ref{table:bptt_arch_gen} shows that initialization with random real images consistently outperforms Xavier initialization. Since the choice of initialization is a very minor design choice and initialization with random real images produces better distilled data, we decided to study BPTT distilled data that was initialized with random real images.

\begin{table}[]
\centering
\begin{tabular}{@{}lcc@{}}
\toprule
          & Xavier Initialization & Real Image Initialization \\ \midrule
AlexNet \cite{krizhevsky2012imagenet}   & 18.13 \%               & 34.66\%                   \\
VGG    \cite{simonyan2015very}   & 25.27  \%               & 37.54\%                      \\
VGG-19   \cite{simonyan2015very} & 22.56 \%               & 32.53\%                   \\
ResNet-34 \cite{he2016deep} & 18.8\%               & 25.58\%                   \\
ViT   \cite{dosovitskiy2020image}    & 20.63 \%               & 25.94\%                   \\ \bottomrule
\end{tabular}
\caption{\textbf{Cross-architecture generalization of BPTT distilled images with different initialization.} The best classification accuracy on five different architectures from 1000 iterations of random hyperparameter search on learning rate, momentum, weight decay, SGD vs. Adam \cite{kingma2014adam, loshchilov2018decoupled}, training iteration, and whether or not to gradient clip to performed to obtain the best classification accuracy. }
\label{table:bptt_arch_gen}
\end{table}

\end{document}